\documentclass[10pt,twocolumn,letterpaper]{article}

\usepackage{cvpr}              %

\usepackage[accsupp]{axessibility}

\definecolor{cvprblue}{rgb}{0.21,0.49,0.74}
\usepackage[pagebackref,breaklinks,colorlinks,allcolors=cvprblue]{hyperref}

\usepackage{graphicx}
\usepackage{booktabs}
\usepackage{multirow}
\usepackage{enumitem}
\usepackage{pifont}%
\usepackage{afterpage}
\usepackage[table]{xcolor}%
\usepackage{etoolbox}  %
\usepackage{mathtools}

\definecolor{static}{HTML}{8ACACA}
\definecolor{unstable}{HTML}{D46F6F}
\definecolor{stable}{HTML}{B8943E}
\definecolor{babyblue}{rgb}{0.54, 0.81, 0.94}
	\definecolor{yellowbanana}{rgb}{1.0, 0.88, 0.21}
 \definecolor{carrotorange}{rgb}{1.0, 0.65, 0.0}
\definecolor{tabfirst}{rgb}{0.5, 1.0, 0.5} 
\definecolor{tabsecond}{rgb}{1, 1, 0.5}  

\newcommand{\cmark}{\ding{51}}%
\newcommand{\xmark}{\ding{55}}%

\newcommand{\Indexed}[1]{#1^{n}}
\newcommand{\Setof}[1]{\{#1^{n}\}}
\newcommand{\HAMER}{HaMeR\xspace}
\def\methfull{\underline{C}onstrained \underline{O}ptimisation and \underline{P}ropagation\xspace}
\def\meth{COP\xspace}

\def\Static{\textcolor{static}{Static}\xspace}
\def\UnstableContact{\textcolor{unstable}{Unstable Contact}\xspace}
\def\StableGrasp{\textcolor{stable}{Stable Grasp}\xspace}

\def\taskfull{\underline{R}econstructing \underline{O}bjects  along \underline{H}and \underline{I}nteraction \underline{T}imelines\xspace}
\def\task{ROHIT\xspace}

\def\timelinefull{\underline{H}and \underline{I}nteraction \underline{T}imeline\xspace}
\def\timeline{HIT\xspace}

\def\paperID{***} %
\def\paperID{7} %
\def\confName{CVPR}
\def\confYear{2026}

\title{Reconstructing Objects along Hand Interaction Timelines in Egocentric Video}

\author{
Zhifan Zhu$^{1}$ \quad
Siddhant Bansal$^{1}$ \quad
Shashank Tripathi$^{2}$ \quad
Dima Damen$^{1}$\\[4pt]
$^{1}$University of Bristol, UK
\quad$^{2}$Max Planck Institute for Intelligent Systems, Tübingen, Germany\\
\url{https://zhifanzhu.github.io/objects-along-hit}\\[4pt]
}

\begin{document}
\maketitle

\begin{abstract}

    We introduce the task of \taskfull (\task). %
    We first define the~\timelinefull (\timeline) from a rigid object's perspective. 
    In a \timeline, an object is first static relative to the scene, then is held in hand following contact, where its pose changes. This is usually followed by a firm grip during use, before it is released to be static again \wrt to the scene. 
    We model these pose constraints over the \timeline, and propose to propagate the object's pose along the \timeline enabling superior reconstruction using our proposed \methfull (\meth) framework.
    Importantly, we focus on timelines with stable grasps -- i.e. where the hand is stably holding an object, effectively maintaining constant contact during use. This allows us to efficiently annotate, study, and evaluate object reconstruction in videos without 3D ground truth.

   We evaluate our proposed task, \task, over two egocentric datasets, HOT3D and in-the-wild EPIC-Kitchens.
   In HOT3D, we curate 1.2K clips of stable grasps. In EPIC-Kitchens, we annotate 2.4K clips of stable grasps including
    390 object instances across 9 categories from videos of daily interactions in 141 environments.
    Without 3D ground truth, we utilise 2D projection error to assess the reconstruction.
    Quantitatively, \meth improves stable grasp reconstruction by $6.2-11.3\%$ and \timeline reconstruction by up to $24.5\%$ with constrained pose propagation.
    
\end{abstract}

\section{Introduction}
\label{sec:intro}

Accurately reconstructing three-dimensional hand-object interactions is key to unlocking many perception problems, including fine-grained understanding of interactions, but also potential applications in augmented reality, robotic imitation learning and human-machine interactions.

\begin{figure}[t]
    \centering
    \includegraphics[width=\linewidth]{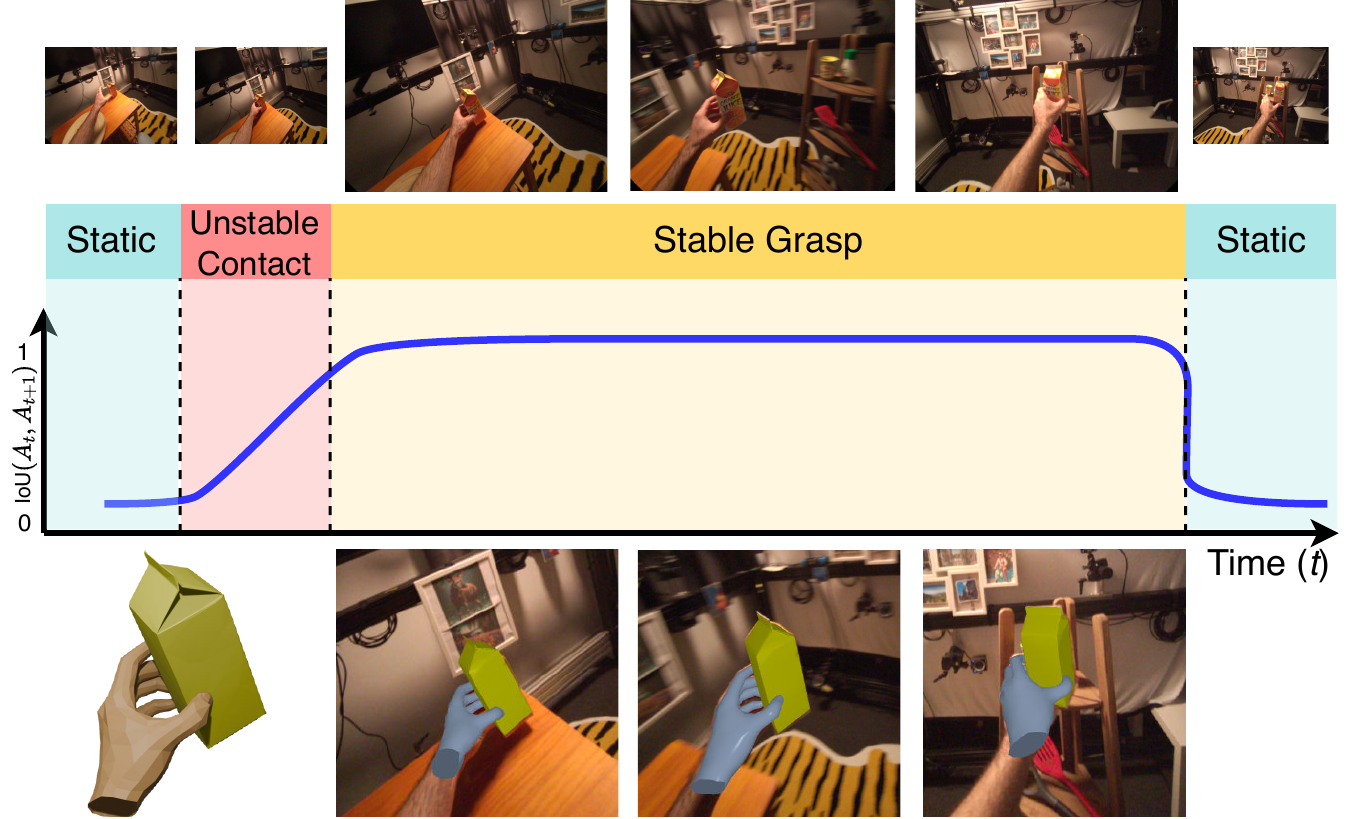}
    \caption{
      Sample HIT sequence from HOT3D~\cite{banerjee2025hot3d} with reconstruction results by our method.
      We illustrate the three types of temporal segments in hand-object interactions: \Static: where the object is static relative to the scene, \UnstableContact: where the hand is firming its grip on the object; and \StableGrasp: where hand is securely holding the object stably, until it is \Static again when put down.
    The plot illustrates the IoU of the in-contact vertices across neighbour frames; for formal definition, refer to~\cref{sec:hoi_timeline_definition}.
    \vspace*{-6pt}
    }
    \label{fig:teaser}
\end{figure}

\begin{figure*}[!t]
    \centering
    \includegraphics[width=\linewidth]{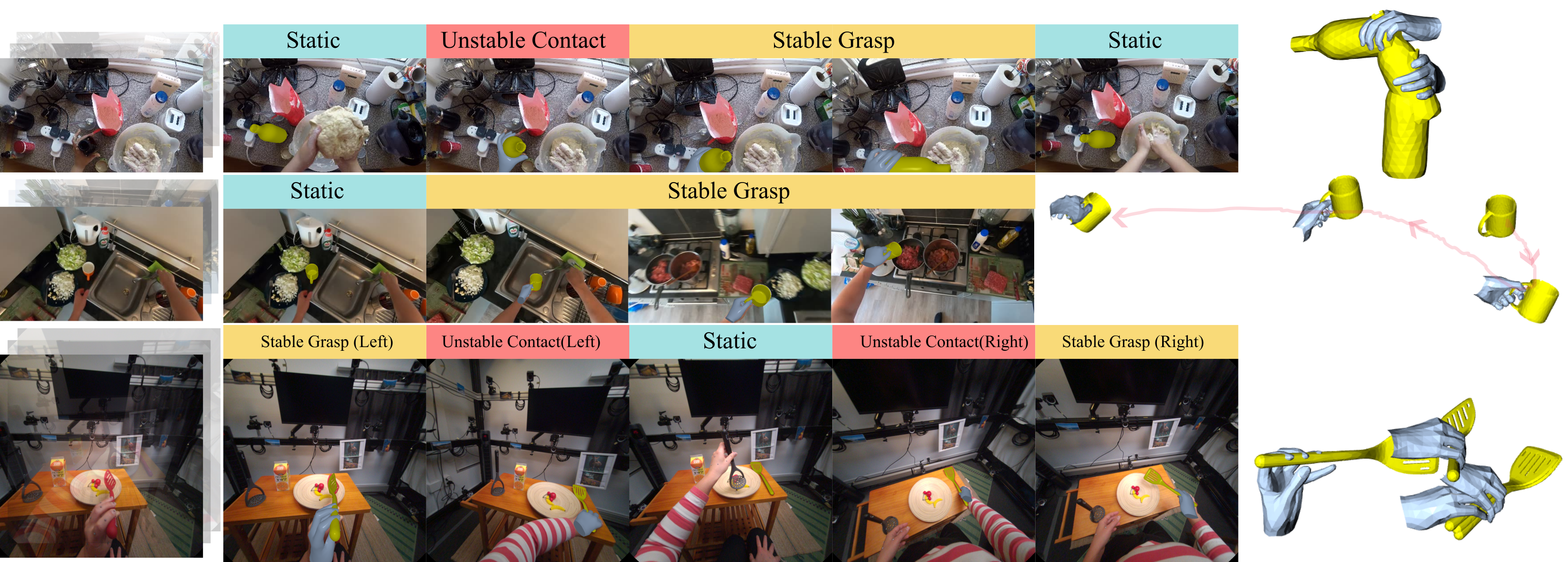}
\caption{\textbf{Qualitative results.} 
    Given a \timelinefull (\timeline) - with an object in \Static, \UnstableContact and \StableGrasp interaction segments, our method, \meth, reconstructs hand (blue) and object (yellow) meshes along the \timeline.
    We show input frames (left), projected meshes (middle) and meshes in 3D world coordinate system (right). Rows 1-2 from EPIC-HIT and row 3 from HOT3D-HIT.%
    }
    \vspace*{-12pt}
    \label{fig:qual_teaser}
\end{figure*}

Efforts to reconstruct the object, in-hand, have typically used 3D ground-truth supervision, paired with 3D~\cite{TaheriGRAB:Objects, BrahmbhattContactPose:Pose, Zhang2021ManipNet:Representation}
or 2D input~\cite{YeWhatsHands, Chen2022AlignSDF:Reconstruction, Hasson2021TowardsVideos, Karunratanakul2020GraspingGrasps, Yang2020CPF:Interaction}.
While such approaches are typically developed to be view-independent, early insights~\cite{Hasson2021TowardsVideos, CaoReconstructingWild, Patel2022LearningVideos} demonstrate that egocentric footage remains significantly challenging, partly due to the considerable occlusion of the object by the hand.

In this work, we particularly focus on egocentric videos, and \textbf{for the first time} consider the complete timeline of the interaction -- before contact, when in-hand, and after release.
We model interactions with \textit{functional intent} -- i.e. where the object is operated or moved securely, rather than simply poked or touched.
We show that during these functional grasps, the same hand and object vertices remain in contact but the object's pose still changes relative to the hand due to finger and hand articulations. 

To estimate the object pose along this timeline, we propose constraints that depend on the type of interaction segment (\Static, \UnstableContact or \StableGrasp) as shown in \cref{fig:teaser}. 
We propose the \methfull (\meth) framework to optimise the object pose during each interaction segment, then propagate the pose to initialise the next segment.
We show sample reconstructions along the \timeline in~\cref{fig:qual_teaser}.

\noindent To summarise, our contributions are as follows:

\begin{itemize}
    \item We propose the task of \taskfull (\textbf{\task}) to reconstruct 3D poses, for rigid known objects, in egocentric videos.
    \item We propose the \methfull (\textbf{\meth}) framework that 
    optimises the object pose for various interaction segment types. 
    \item We curate hand interaction timelines from the  egocentric HOT3D dataset, to evaluate our method with 3D ground truth. We refer to this as \textbf{HOT3D-\timeline} dataset.
    \item We label $2.4K$ stable grasps clips from EPIC-Kitchens, along with $96$ \timeline, which we call the \textbf{EPIC-\timeline} dataset.
    \item We evaluate \meth on both datasets and show $6.2-11.3\%$ improvement in stable grasp reconstruction and up to $24.5\%$ gain in \timeline reconstruction with propagation.
\end{itemize}

\section{Related Works}
\label{sec:related}

Here we discuss work on hand pose estimation, object pose estimation and joint hand-object reconstruction. For comparison of hand object reconstruction datasets, see~\cref{tab:datasetCompare}.

\noindent \textbf{3D Hand Pose Estimation.}
Estimating 3D hand pose from images has been proposed for both free hands and hands in-interactions.
FrankMocap~\cite{RongFrankMocap:Integration} is a commonly used CNN-based model in many hand-object reconstruction methods~\cite{Hasson2021TowardsVideos, CaoReconstructingWild, Patel2022LearningVideos, YeWhatsHands, ye2023vhoi}.
METRO~\cite{lin2021end-to-end} proposes to use a transformer on top of the CNN feature for regression.
WildHands~\cite{Prakash2024Hands} addresses the perspective distortion for egocentric hand pose estimation.
Recently, fully transformer-based methods~\cite{pavlakos2024reconstructing, dong2024hambasingleview3dhand, zhou2024simple, potamias2025wilor, li2024hhmr} train on scaled training data with higher network capacity, they achieve superior performance.
The transformer based \HAMER~\cite{pavlakos2024reconstructing} has been recently used as guidance for full body estimation task~\cite{yi2024egoallo} to show its usefulness.
In this work, our in-the-wild pipeline also uses \HAMER~\cite{pavlakos2024reconstructing}  due to its robust performance. 

\noindent \textbf{3D Object Pose Estimation.}
A full review of works on estimating 3D objects pose from single image is out of our scope. Here, we focus on relevance for hand-object interaction scenarios. To estimate 3D object pose, several works~\cite{chen2021monorun, ornek2024foundpose, goodwin2022zero, liu2021utoshape, wang2019normalized} assume a known object shape template and estimate 6-DoF pose by fitting it to 2D image features (e.g. masks or keypoints), using either 2D-3D~\cite{chen2021monorun, ornek2024foundpose} or 3D-3D correspondences~\cite{goodwin2022zero, liu2021utoshape, wang2019normalized}. 
Other methods~\cite{meshrcnn, yu2018posecnn} estimate object pose and shape jointly, however, at the cost of geometric fidelity.  
While effective, these works typically assume unoccluded objects. %

\noindent \textbf{3D Hand-Object Reconstruction.}
Methods are grouped into two categories.
The first category, known-CAD methods, assumes that object CAD models are given and fits 3D shapes into 2D observations. 
These can further be classified into 
data-driven~\cite{kwon2021h2o,lin2023harmonious, LiuSemi-SupervisedTime, Yang2022Artiboost,Wang2023Interacting, Tse2022Collaborative, Yang2020CPF:Interaction, Aboukhadra2023Thor,cho2023transformer,ismayilzada2025qort} or
optimisation-based~\cite{Hasson2021TowardsVideos, CaoReconstructingWild, Patel2022LearningVideos} methods.
Data-driven methods learn to jointly reconstruct hands and objects from seen object examples, 
whereas optimisation-based methods address the reconstruction by directly fitting to 2D signals.
RHO~\cite{CaoReconstructingWild} is the first optimisation based single-frame method.
The optimisation-based methods~\cite{CaoReconstructingWild, Hasson2021TowardsVideos, Patel2022LearningVideos} share the same pipeline where hand/object is first independently optimised, followed by joint optimisation with physical constraint terms.
HOMan~\cite{Hasson2021TowardsVideos} generalises to multiple frames and includes temporal smoothness of mesh vertices over time.

The second category, CAD-agnostic methods, estimates the object pose without using explicit CAD models. 
Many CAD-agnostic methods~\cite{HassonLearningObjects,Karunratanakul2020GraspingGrasps,Chen2022AlignSDF:Reconstruction,YeWhatsHands,ye2023ghop, Prakash2024HOI, chen2025hort,yu2025dynamic} learn object shape priors,
 or retrieve from (generative-)object pools~\cite{jiang2025hand,liu2025easyhoi,aytekin2025follow}. 
The other line of methods~\cite{Huang2022ReconstructingVideo,hampali2023inhand,fan2024hold, wang2025magichoi} uses neural networks to fit the underlying object shape from multiple views. 
We experimentally show that CAD-agnostic methods are incapable of generalising to in-the-wild egocentric videos.

Our method belongs to the first category, adopting a simplified assumption needed for the challenges of in-the-wild reconstruction.
This also allows us to focus on \timeline reconstruction, which has not been attempted before.
Different from all previous optimisation-based methods, 
we examine the object's relative motion through various segments and reconstruct it throughout the \timeline.

\section{The \task Task}
\label{sec:problem_def}

\subsection{What is a Hand Interaction Timeline?}
\label{sec:hoi_timeline_definition}

A \timelinefull (\timeline) for a given object is a video sequence that focuses on the hand interacting with one object to either use or move that object. From the object's perspective, the \timeline can be divided into many contiguous segments of three types: 
\begin{itemize}
    \item \Static: before/after a hand interaction, the object is typically supported by a surface and is thus static relative to the scene/world.
    \item \StableGrasp: the hand grasps the object firmly, allowing functional usage and secure movement (details below).
    \item \UnstableContact: the object is neither Static nor in Stable Grasp, \eg when a grip is being formed on the object.
\end{itemize}
As in Fig~\ref{fig:qual_teaser}, the \timeline can involve many temporal segments (e.g. \Static $\rightarrow$ \UnstableContact $\rightarrow$ \StableGrasp $\rightarrow$ \UnstableContact $\rightarrow$ \StableGrasp $\rightarrow$ \Static).
Our task is to reconstruct the hand-object interaction along the full \timeline.

The term \StableGrasp has been previously used in human grasp analysis~\cite{Bullock2013HandCentric, Cutkosky1989OnGraspChoice, Feix2016TheTypes}.
While definitions vary, they centre around the object being ``held securely with one hand, irrespective of the hand orientation''~\cite{Feix2016TheTypes}.  
Intuitively, this means that \textit{the same hand and object vertices remain in contact for the grasp duration}.

\begin{figure}[t]
    \centering
    \includegraphics[width=\linewidth]{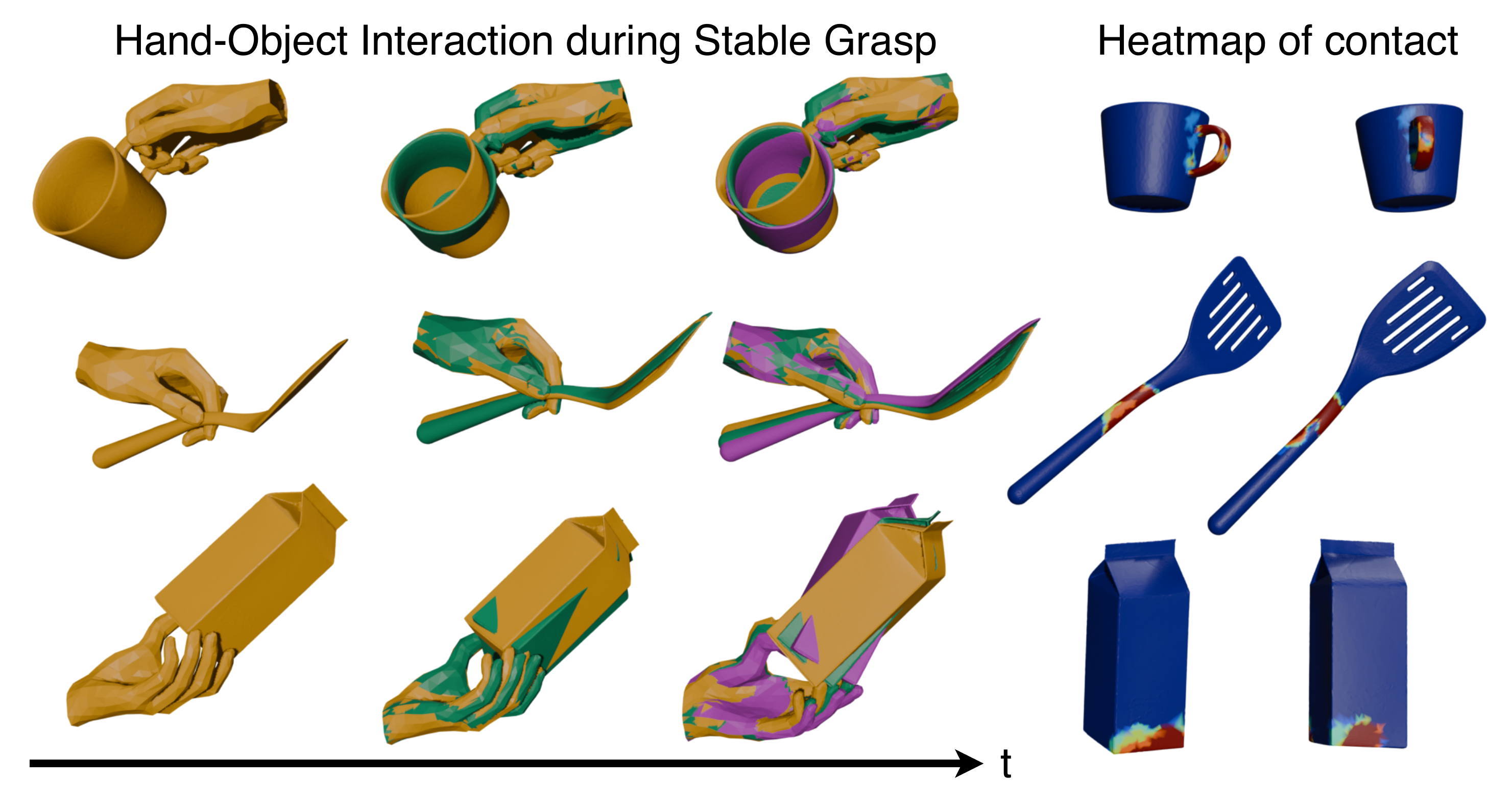}
    \vspace*{-16pt}
    \caption{\textbf{Stable Grasp Intuition.} 
    Three samples from HOT3D.
    In each row, we align the hand coordinate system for three frames from one stable grasp.
    Left: finger articulations and object pose vary over time.
    Right: contact area (shown as a heat map of objects vertices in contact with the hand) remains consistent.
    }
    \vspace*{-6pt}
    \label{fig:visual_grasp_def}
\end{figure}

\begin{figure*}[!t]
\centering
\includegraphics[width=\linewidth]{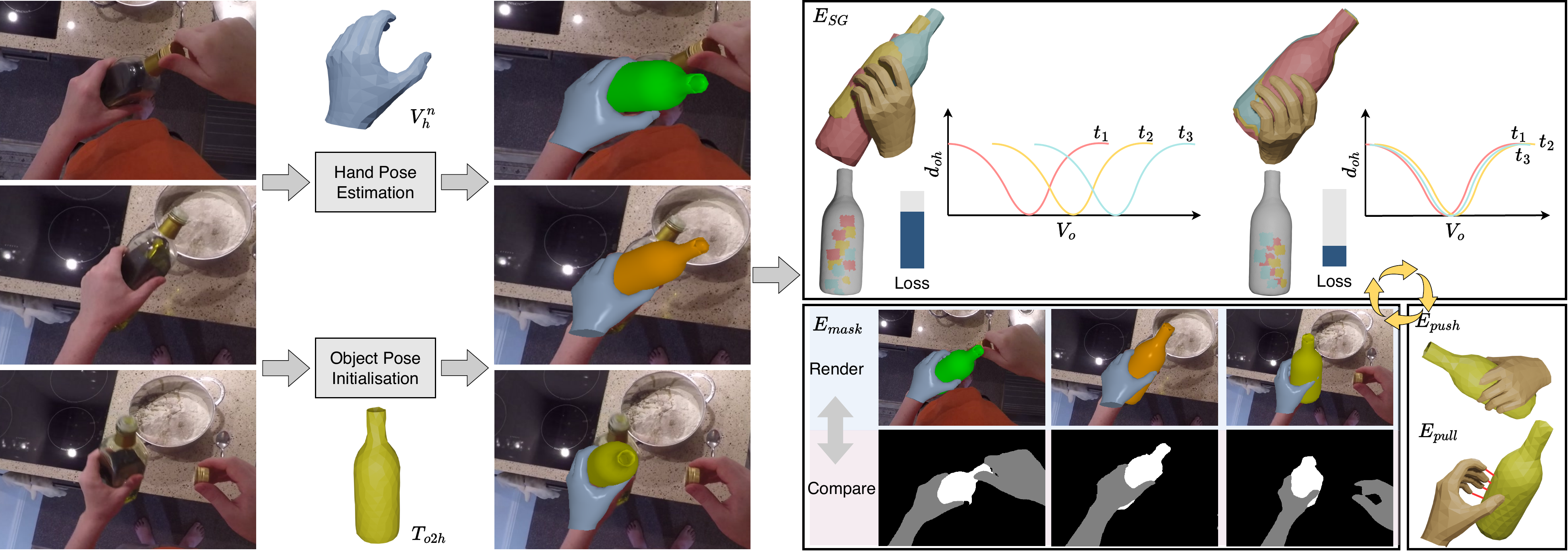}
\caption{
\textbf{Optimising the \StableGrasp segment}. 
We show three frames within one stable grasp.
We utilise \HAMER~\cite{pavlakos2024reconstructing} to reconstruct the hand mesh in-the-wild.
We initialise $\Indexed{T_{o2h}}$ to one $T_{o2h}$, but keep the diverse finger articulations from the hand pose estimates.
During optimisation, we measure the distance between each hand vertex and object vertices $V_o$. In the left plot, we show that the contact area $d_{oh} \approx 0$ differs over time (visualised on the gray bottle).
The novel loss, $E_{SG}$, 
minimises the variation of distance between the hand and object vertices over time, by adjusting the object's pose relative to the hand.
As $E_{SG}$ is minimised, the contact area is aligned (see updated plot).
Additional losses are used to regularise the optimisation:
$E_{mask}$ renders the reconstruction and compares it against estimated object masks while $E_{push}$ and $E_{pull}$ respectively ensure the object is not penetrated by or away from the hand.
}
\vspace*{-12pt}
\label{fig:method}
\end{figure*}

Formally, for any pair of frames $i$ and $j$ within the \StableGrasp, 
we use $S_i$ and $S_j$ to denote the in-contact area on the object surface,
and intersection-over-union $\mathrm{IOU}(S_i, S_j)$ between in-contact areas.
Following above intuition, the duration of the stable grasp is defined as:
\begin{equation}
\begin{split}
\label{eq:auto_grasp_algo}
    \left[l^*, r^*\right] = \underset{l, r}{\mathrm{argmax}} \left(r - l \right) \quad \\
    \textrm{s.t.} \; \mathrm{IOU}(S_i, S_j) > \tau \;\; \quad \forall l \leq i < j \leq r 
\end{split}
\end{equation}
where $\tau$ specifies the minimum IOU threshold. 
The $\mathrm{argmax} \left(r-l\right)$ implies the duration of the stable grasp. 
Importantly, while there is a \textit{consistent contact area}, the hand orientation and finger articulations/poses vary during the stable grasp. This allows the object pose to vary relative to the hand -- see~\cref{fig:visual_grasp_def}.

Similarly if $S_i = \emptyset$, the object is not in contact with the hand and is thus assumed \Static.
If $S_i \neq \emptyset$ but $\mathrm{IOU}(S_i, S_j) \le \tau$, the object is in an \UnstableContact.

\subsection{\task task and Notations}
\label{sec:notations}

\noindent\textbf{Assumptions.} \task aims to jointly estimate the object poses  over hand interaction alone one \timelinefull (\timeline). 
This allows reconstructing 3D mesh of the hand and the object, per frame.
We assume the knowledge of object category CAD model,
as well as the hand-side (left/right) associated with each segment.
We also assume the temporal boundaries within the \timeline, \ie the start/end of each segment.
Note that the assumption of hand side and in-hand segment boundary is implicit in \emph{all prior works} that assume the hand is already grasping the object. 
These annotations expect the durations (start-end) of the hand grasps as input.
Relaxing the assumptions is discussed in the supp.

Following prior works~\cite{Hasson2021TowardsVideos,CaoReconstructingWild,Patel2022LearningVideos}, we use MANO~\cite{Romero2017EmbodiedTogether} to represent the \textbf{hand mesh}, 
which takes as input the per-frame finger articulation vector $\theta^n \in \mathbb{R}^{45}$ 
and outputs the hand mesh with vertices $\Indexed{V_{h}} = \mathrm{MANO}(\Indexed{\theta}) \in \mathbb{R}^{778 \times 3}$ in the \textit{hand coordinate system}. 
When unavailable, we utilise an off-the-shelf method~\cite{pavlakos2024reconstructing} to obtain finger articulations for each frame $\Indexed{\theta}$.
Additionally, we utilise the hand-to-camera (\textit{h2c}) pose $\Indexed{T_{h2c}}$, which determines the hand wrist orientation and position.
$\Indexed{T_{h2c}}$ is used to transform meshes from the hand coordinate system to the \textit{camera coordinate system} for each frame.

For the \textbf{object mesh}, $V_o \in \mathbb{R}^{|V_o| \times 3}$ denotes the known object vertices in the \textit{object coordinate system}. 
We denote the object-to-hand~($o2h$) poses $\Indexed{T_{o2h}}$ and the object scale $s \in \mathbb{R}$, which transform the object vertices to $\Indexed{V_{o:h}}$ in the \textit{hand coordinate system} for each frame.
Given the hand-to-camera pose $\Indexed{T_{h2c}}$, the object mesh in the camera coordinate is represented as 
\begin{equation}
    \Indexed{V_{o:c}} = \Indexed{T_{h2c}} \, (\Indexed{T_{o2h}} (s * V_o))
\label{eq:vo_via_hand}
\end{equation}

Lastly, to reconstruct the \timeline in world coordinates, 
we use the world-to-camera pose $\Indexed{T_{w2c}}$.
In \Static segments (\cref{sec:optim_static_segment}) this allows us to represent the object mesh in the camera coordinate as
\begin{equation}
    \Indexed{V_{o:c}} = \Indexed{T_{w2c}} (\Indexed{T_{o2w}} (s * V_o))\label{eq:vo_via_world}
\end{equation}
where $\Indexed{T_{o2w}}$ is object-to-world pose.
In \StableGrasp and \UnstableContact segments,
the hand-to-world pose $\Indexed{T_{h2w}}$, 
obtained via $\Indexed{T_{h2w}} = (\Indexed{T_{w2c}})^{-1} \times \Indexed{T_{h2c}}$, 
is used to convert between the hand coordinate and the world coordinate,
\begin{equation}
\label{eq:conversion_hand_world}
    \Indexed{T_{o2w}} \xrightleftharpoons[\Indexed{T_{h2w}}]{(\Indexed{T_{h2w}})^{-1}} \Indexed{T_{o2h}}
\end{equation}
The world-to-camera pose $\Indexed{T_{w2c}} =  (\Indexed{T_{c2w}})^{-1}$ is provided from egocentric datasets~\cite{EPICFields2023, banerjee2025hot3d} or can be estimated~\cite{SchonbergerStructure-from-MotionRevisited}.

\section{\underline{C}onstrained \underline{O}ptimisation and \underline{P}ropagation}
\label{sec:recon_timeline}

We propose the \methfull~(\meth) framework to reconstruct hand-object mesh pairs 
along the \timeline.
Our proposal stems from the understanding of the various constraints governing the changing in-contact vertices of the object along the \timeline (Fig~\ref{fig:teaser}).
To accommodate variable length segments, we always use $N$ sampled frames per segment.
Once each segment is optimised, using specific constraints, the pose is propagated to initialise the next segment -- as object poses have to remain temporally smooth.
We first explain the constrained optimisation for each type of segment.

\subsection{Optimising a \Static Segment}
\label{sec:optim_static_segment}

While the object appears moving in the camera due to head motion in an egocentric video, during the \Static segment, the object is stationary in the world coordinate system.

\noindent \textbf{\Static Constraint.} When the object is static, the object-to-world pose $T_{o2w}$ remains fixed across all $N$ frames in the segment, 
$T_{o2w} = \Indexed{T_{o2w}}$.
We use~\cref{eq:vo_via_world} to get the object mesh in camera coordinate $\Indexed{V_{o:c}}$, and 
optimise $T_{o2w}$ and the scale $s$ using the \textit{render-and-compare} loss. 

\noindent \textbf{Render-and-Compare Loss} ($E_{mask}$)\textbf{:} 
This loss focuses on estimating a reconstruction that best matches the 2D projections of object masks throughout the sequence. 
We measure the error via sum of pixel differences: 
\begin{equation}
\label{eq:e_mask}
    \Indexed{E_{mask}} = | \Indexed{\mathcal{C}_o} \otimes (\Indexed{\mathcal{M}_o} - {\Pi(\Indexed{V_{o:c}})})|^2_2
\end{equation}
where $\Indexed{\mathcal{M}_o}$ is the object mask which we use for supervision and $\Pi(\cdot)$ is the differentiable projection function~\cite{KatoNeuralRenderer}.
$\Indexed{\mathcal{C}_o}$ is the occlusion-aware mask as in~\cite{Hasson2021TowardsVideos, ZhangPerceivingWild} 
which only computes the error within regions of the object that are not occluded, %
set to 1 for the object and the background, and 0 for the hand.
This masking avoids penalising the missing parts of the object due to hand occlusion.

\subsection{Optimising a \StableGrasp Segment}
\label{sec:optim_stable_grasp}

In a stable grasp, 
the object's pose is controlled by the in-hand contact vertices.
We thus optimise the object relative pose \wrt the hand, \ie $\Indexed{T_{o2h}}$.
This is different from optimising the object \wrt to the camera~\cite{Hasson2021TowardsVideos,Patel2022LearningVideos}.
We use~\cref{eq:vo_via_hand} to get the object mesh in camera coordinate frame $\Indexed{V_{o:c}}$,
and optimise $\Indexed{T_{o2h}}$ and $s$.

\noindent \textbf{\StableGrasp Constraint.} Following the formal definition from~\ref{sec:hoi_timeline_definition}, we use the stable contact area as our constraint to optimise this segment type. Recall that the object pose relative to the hand can change as long as the vertices in contact with the hand remain stable. We introduce a new loss $E_{SG}$ to model this constraint.

\noindent \textbf{\StableGrasp Loss} ($E_{SG}$)\textbf{:}
First, we limit this to the hand vertices that are typically in contact with objects -- these are the five fingertips (see supp. for visualisation).
More formally, let's denote this subset of hand vertices as \( V_F \subset V_h \). 
For each object vertex \( v_o \in V_o \), 
we calculate the distance \( d_{oh} \) to each \( v_h \in V_F \). 
We then minimize the average variation of this distance across all pairs of frames \( n \) and \( m \):
\begin{align}
\label{eq:e_SG}
E_{SG} & = \sum_{v_o \in V_o} \sum_{v_h \in V_F} \sum_{n=1}^N \sum_{m=1}^N
 |d_{oh}^n - d_{oh}^m|_1 \\
 \label{eq:e_SG2}
d_{oh}^n & := |v_o^n - v_h^n|_2^2
\end{align}
where \( v_h \) refers to one hand vertex, with \( v_h^n \) representing its location at frame \( n \); similarly, \( v_o \) and \( v_o^n \) denote the object vertex and its frame-specific location, respectively.

Note that we assume a rigid object, so we can only minimise $E_{SG}$ by updating its overall pose -- i.e. translating and rotating the object relative to the hand.
As $E_{SG}$ is minimised, the object's pose is optimised so as to minimise the difference between $d_{oh}^n$ and $d_{oh}^m$ for all pairs of frames.
Optimising for this distance is the same as aligning the contact area -- \ie if two frames have the same hand-object vertex distance, then the contact area will undoubtedly be aligned.

\Cref{fig:method} visualises this loss by 
considering three timestamps for the hand and bottle ($t_1$, $t_2$, $t_3$).
We consider a single hand vertex and plot the distance to all object vertices over time.
Before optimising the $E_{SG}$ loss, the distances to bottle vertices changes over time -- visualised through the $d_{oh}$ coloured curves (left graph).
After optimising for $E_{SG}$, the plots are better aligned (right graph).
We visualise the contact area on the bottle before/after minimising $E_{SG}$.

\begin{table*}[t]
\centering
    \caption{\textbf{Dataset Comparison}. Here we compare various characteristics and labels provided by various datasets. We also show statistics of \StableGrasp and \timeline (when available). $^\dagger$: subjects in the released train/val set}
    \vspace{-10pt}
\resizebox{1\textwidth}{!}{%
\begin{tabular}{@{}lrccccccccccccccc@{}}\toprule
\multirow{3}{*}{Dataset} & \multirow{3}{*}{Year} &\multicolumn{3}{c}{\textbf{Characteristics}} &\multicolumn{3}{c}{\textbf{Labels}} &\multicolumn{5}{c}{\textbf{Stable Grasps' Stats}} &\multicolumn{4}{c}{\textbf{\timeline's Stats}} \\
\cmidrule(lr){3-5}\cmidrule(lr){6-8}\cmidrule(lr){9-13}\cmidrule(lr){14-17}
& & \small{In-the-wild} & \parbox{0.7cm}{\small{Funct.}\\Intent} & Ego & Pose GT & \parbox{1cm}{Stable \\ Grasp} & \parbox{1cm}{\timeline} & \#Env & \#Sub & \#Cat & \#Inst & \#Seq & \parbox{1cm}{Avg. \\ Duration} & \#frames & \parbox{1.5cm}{Avg. Seg. \\ Per \timeline} & \#Seq \\
 \hline
 HOI4D~\cite{Liu2022HOI4D:Interaction} &2022 & \xmark & \cmark & \cmark & 3D & \xmark & \xmark & 610 & 9 & 20 & 800 & 5,000 & - & - & - & - \\
 ARCTIC~\cite{fan2023arctic} &2023 & \xmark & \cmark & \cmark & 3D & \xmark & \xmark & 1 & 9$^\dagger$ & 11 & 11 & 339 & - & - & - & - \\
 HOGraspNet~\cite{2024graspnet} &2024 & \xmark & \xmark & \cmark & 3D & \cmark & \xmark & 1 & 99 & 30 & 30 & $\sim$3861 & - & - & - & - \\
 HOT3D~\cite{banerjee2025hot3d} &2024 & \xmark & \cmark & \cmark & 3D & \xmark & \xmark & 4 & 19 & 33 & 33 & 295 & - & - & - & - \\
 \textbf{HOT3D-\timeline} (ours) &2025 & \xmark & \cmark & \cmark & 3D & \cmark & \cmark & 4 & 9 & 22 & 22 & 1,239 & 121.1s & 410,650 & 29.1 & 113 \\
 \midrule
 MOW~\cite{CaoReconstructingWild, Patel2022LearningVideos} &2021 & \cmark & \cmark & \xmark & \xmark & \xmark & \xmark & 500 & 500 & 121 & 500 & 500 & - & - & - & - \\ 
 \textbf{EPIC-\timeline} (ours) &2025 & \cmark & \cmark & \cmark & 2D Mask & \cmark & \cmark & 141 & 31 & 9 & $\sim$390 & 2,431 & 13.8s & 79,736 & 2.8 & 96 \\
 \bottomrule
\end{tabular}
}
\vspace*{-12pt}
    \label{tab:datasetCompare}
\end{table*}

In addition to this newly introduced loss, we also use %
$E_{mask}$, introduced in Eq~\ref{eq:e_mask}, as in the \Static segment enforces the reconstruction to match the observation of the object in the images.
We also use two standard physical heuristic losses $E_{push}$ and $E_{pull}$ to ensure contact and avoid mesh penetration between the hand and the object.

\noindent \textbf{Push} ($E_{push})$ \textbf{and Pull }($E_{pull}$)\textbf{ Loss :}
Motivated by previous works~\cite{Yang2020CPF:Interaction,CaoReconstructingWild,Hasson2021TowardsVideos,Patel2022LearningVideos},  $E_{push}$ pushes the object out of the penetrating region against the hand while the balancing loss $E_{pull}$ pulls the object to touch the hand. 
For calculations of $E_{push}$ and $E_{pull}$, refer to the supp.

Combining the four losses, the objective function for the \StableGrasp segment becomes:
\begin{equation}
\label{eq:totError}
    \resizebox{\linewidth}{!}{$
    E(\Setof{T_{o2h}}, s) =
     \lambda_1 E_{SG} + \sum\limits_{n=1}^N  (\Indexed{E_{mask}} + \lambda_2\Indexed{E_{push}} + \lambda_2\Indexed{E_{pull}} )
    $}    
\end{equation}
where $\lambda_1$ is the weight for $E_{SG}$ and $\lambda_2$ is the weight for $\Indexed{E_{push}}$ and $\Indexed{E_{pull}}$ and $s$ is the object scale. 

The stable grasp optimisation is overviewed in~\cref{fig:method}. 
$E_{SG}$ is key to optimising a Stable Grasp as it optimises jointly across frames.
Note that prior work~\cite{Hasson2021TowardsVideos} does add temporal smoothing over time, which we experimentally show to be insufficient for accurate optimisation.

\subsection{Optimising an \UnstableContact Segment}
\label{sec:optim_unstable_contact}

When the object is beyond the stable grasp, but still in the hand, 
we only make assumption on the contact with the hand.
We thus utilise $E_{mask}$, $E_{push}$, $E_{pull}$ losses \textit{without} the stable grasp assumption, \ie set $\lambda_1 = 0$ in~\cref{eq:totError}.

\subsection{Pose \underline{P}ropagation in \meth}
\label{sec:propagation}

\begin{figure}
    \centering
    \includegraphics[width=\linewidth]{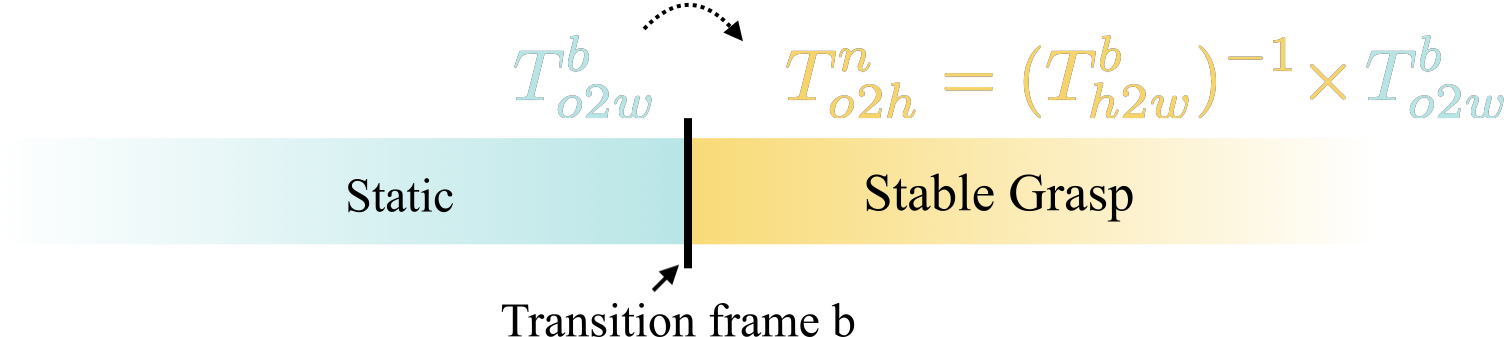}
    \caption{Sample propagation (e.g. from \Static to \StableGrasp). }
    \vspace*{-12pt}
    \label{fig:propagation_at_transition}
\end{figure}

When we optimise along the \timeline, each segment is optimised in order.
\Cref{fig:propagation_at_transition} illustrates an example propagation from \Static to \StableGrasp segment.
Once the segment is optimised, 
we obtain the object-to-world pose at the last frame, $b$, which is the transitioning pose to the next segment $T^b_{o2w}$. 
Intuitively, the object-to-world pose should be consistent at the transition frame $b$.
Where needed,~\cref{eq:conversion_hand_world} is utilised to convert from the hand coordinate into the world coordinate, or back from the world to the hand coordinate.

The optimal pose at $b$ is passed to the next consecutive segment as an additional initialisation.
Propagating the object-to-world pose at the transition is key to our proposed propagation.
Notice that both the pose of the object and its scale are used to initialise the next segment.
We propagate between \textbf{all consecutive segments} regardless of the type, \eg \StableGrasp $\rightarrow$ \Static or \Static $\rightarrow$ \UnstableContact.

Each initialisation, including the propagated one, is optimised independently based on the constrained losses for the segment type (Sec~\ref{sec:optim_static_segment}-\ref{sec:optim_unstable_contact}). 
We select the pose with the minimum $E_{mask}$ as our method's prediction.
This is again passed to the next segment in the \timeline, and so on.

\section{Experiments}
\label{sec:experiments}

\subsection{Dataset}
\label{subsec:dataset}
With the definition of \timeline and \StableGrasp in ~\cref{sec:hoi_timeline_definition}, we annotate timelines from unscripted egocentric videos.
In HOT3D~\cite{banerjee2025hot3d}, we automatically annotate HITs from the 3D ground truth -- we refer to these sequences as \textbf{HOT3D-\timeline}.
Object masks are provided by the ground truth.
In EPIC-KITCHENS~\cite{DamenScalingDataset}, we manually annotate \timeline segments for 9 rigid and commonly used object categories (plate, bowl, bottle, cup, mug, can, pan, saucepan, glass).
Corresponding object masks are available from the VISOR dataset~\cite{VISOR2022}.
We refer to this datasets as \textbf{EPIC-\timeline}.

We compare against existing datasets in hand-object reconstruction in~\cref{tab:datasetCompare} (full table in supp.).
For HOT3D, we automatically extract 1,239 stable grasps and extend them to 113 HITs covering 3,288 segments (w/ 872 \Static and 1177 \UnstableContact).
For EPIC-KITCHENS, we labelled 2,431 video clips of \StableGrasp from 141 distinct videos in 31 kitchens.
We additionally manually label 96 HITs, covering 296 segments (135 \Static, 106 \StableGrasp, and 28 \UnstableContact).
Details in supp.

\subsection{Implementation Details}
\label{sec:implementation}

Following HOMan~\cite{Hasson2021TowardsVideos}, we sparsely and linearly sample 30 frames from each segment for optimisation and evaluation.
For each object, we use 10 rotation initialisations and 1 global translation to initialise $T_{o2h}$,
following~\cite{Hasson2021TowardsVideos}.
The error $E_{mask}$ is defined in pixels whilst $E_{SG}$, $E_{push}$ and $E_{pull}$ are defined in 3D space (metres).
We use the camera's focal length $f$ as a scaling factor: ${\lambda_f = f * \texttt{render\_size}}$.
We use $\texttt{render\_size} = 256$ and set $\lambda_1 = \lambda_f$ and $\lambda_2 = 0.1 * \lambda_f$ (Eq~\ref{eq:totError}).
The optimisation takes on average 30s on a RTX 4090 for one 30-frame segment.
Please refer to the supp. for additional implementation details.

\subsection{Baselines and Quantitative Metrics}
\label{sec:define_metrics}

We focus on baselines that are able to reconstruct object poses in the hand,
using a predefined CAD model. 
Since \meth does not require training, we do not compare with data-driven methods~\cite{lin2023harmonious, fan2023arctic,lin2023harmonious, Yang2020CPF:Interaction, Yang2022Artiboost}. 
We compare to:
\begin{itemize}[leftmargin=5mm,itemsep=-1.5ex,partopsep=1ex,parsep=2ex]
\item \textbf{HOMan}~\cite{Hasson2021TowardsVideos}---A common CAD-based baseline that progressively optimises the object pose relative to the hand. HOMan implements temporal smoothing and uses $E_{pull}$ and $E_{push}$.
Similar to our method, we use VISOR~\cite{VISOR2022} masks for fair comparison.
\item \textbf{Rigid}~\cite{swamy2023showme}---assumes that objects are not allowed any motion within grasp, minimising overall rotation and translation of the object relative to the hand.
\item \textbf{Dynamic}---A variation of \meth without the stable grasp constraint ($\lambda_1 = 0$).
\end{itemize}
We note that RHOV~\cite{Patel2022LearningVideos} is a relevant baseline but no code is available to be used as a baseline. 
CAD-agnostic methods~\cite{fan2024hold, YeWhatsHands, ye2023vhoi, wu2024reconstructing, Prakash2024HOI} fail catastrophically on our datasets (see supp) making them unsuitable as baselines.

We use two standard metrics for evaluation with/without 3D ground truth, suitable for all segment types:

\noindent \textbf{Average Distance (ADD)} is the standard metric for methods with 3D ground truth. Following~\cite{krull2015learning, yu2018posecnn, Hodan2020Epos, 2024graspnet}, we measure the distance of corresponding vertices between GT and predicted object vertices, and average it over vertices and frames. ADD is 1 for a sequence if average distance is less than 10\% of the object's diameter, and 0 otherwise.
In symmetric CAD models, we calculate the minimal average distance among the valid symmetric transformations~\cite{hodavn2020bop}. 

\noindent \textbf{Intersection-over-Union (IOU)}. 
We use IOU as a proxy of pose accuracy when 3D GT is \textit{not} available. We measure IOU between ground truth mask and rendered mask for the object in camera view. We report average IOU across all frames. In case of occlusion with other components, only the non-occluded area of the rendered projection is used.

We also propose variations of these standard metrics that particularly measure the \StableGrasp in \timeline:

\begin{table*}[!t]
    \caption{\textbf{Results on \StableGrasp in HOT3D-HIT.} 
    \colorbox{green!25}{Green} for \colorbox{green!25}{best} and \colorbox{yellow!25}{yellow} for \colorbox{yellow!25}{second} best.
    $\text{COP}^{\dagger}$ is COP w/o propagation.
    }
    \vspace*{-12pt}
    \centering
    \resizebox{\textwidth}{!}{%
    \setlength{\tabcolsep}{10pt} %
\begin{tabular}{@{}lcccccccccccc}
\toprule
\multirow{2}{*}{Category} & \multicolumn{4}{c}{SCA-IOU} & \multicolumn{4}{c}{ADD} & \multicolumn{4}{c}{SCA-ADD} \\
\cmidrule(lr){2-5}\cmidrule(lr){6-9}\cmidrule(lr){10-13}
 & HOMan~\cite{Hasson2021TowardsVideos} & Rigid~\cite{swamy2023showme} & Dynamic & $\text{COP}^{\dagger}$ & HOMan~\cite{Hasson2021TowardsVideos} & Rigid~\cite{Hasson2021TowardsVideos} & Dynamic & $\text{COP}^{\dagger}$ & HOMan~\cite{Hasson2021TowardsVideos} & Rigid~\cite{swamy2023showme} & Dynamic & $\text{COP}^{\dagger}$ \\
\midrule
bottle\_bbq        & 23.1 & \cellcolor{yellow!25}39.2 & 34.2 & \cellcolor{green!25}43.5 & 4.9  & 30.5  & \cellcolor{yellow!25}52.4 & \cellcolor{green!25}57.3 & 3.8  & 16.9  & \cellcolor{yellow!25}21.0 & \cellcolor{green!25}26.7 \\
bottle\_mustard    & 15.2 & 16.8 & \cellcolor{yellow!25}23.2 & \cellcolor{green!25}32.3 & 0.0  & 12.5  & \cellcolor{yellow!25}37.5 & \cellcolor{green!25}37.5 & 0.0  & 5.9   & \cellcolor{yellow!25}14.9 & \cellcolor{green!25}18.3 \\
bottle\_ranch      & 16.8 & \cellcolor{yellow!25}33.3 & 32.9 & \cellcolor{green!25}40.4 & 9.1  & 31.8  & \cellcolor{yellow!25}50.0 & \cellcolor{green!25}56.8 & 5.9  & 17.4  & \cellcolor{yellow!25}19.6 & \cellcolor{green!25}25.4 \\
bowl               & 39.4 & \cellcolor{green!25}53.4 & 32.8 & \cellcolor{yellow!25}48.7 & 36.5 & \cellcolor{yellow!25}84.9 & 84.9 & \cellcolor{green!25}89.7 & 23.3 & \cellcolor{green!25}44.8 & 28.5 & \cellcolor{yellow!25}43.7 \\
can\_parmesan      & 29.1 & \cellcolor{green!25}44.8 & 32.8 & \cellcolor{yellow!25}43.5 & 0.0  & 19.0  & \cellcolor{yellow!25}27.0 & \cellcolor{green!25}30.2 & 0.0  & 9.0   & \cellcolor{yellow!25}10.4 & \cellcolor{green!25}14.7 \\
can\_soup          & 32.4 & \cellcolor{yellow!25}50.3 & 37.1 & \cellcolor{green!25}50.6 & 3.0  & 14.1  & \cellcolor{yellow!25}21.2 & \cellcolor{green!25}26.3 & 2.5  & 6.7   & \cellcolor{yellow!25}7.6  & \cellcolor{green!25}13.8 \\
can\_tomato\_sauce  & 36.2 & \cellcolor{green!25}46.3 & 33.2 & \cellcolor{yellow!25}45.8 & 2.0  & 2.9   & \cellcolor{yellow!25}9.8  & \cellcolor{green!25}9.8  & 1.6  & 1.1   & \cellcolor{yellow!25}3.2  & \cellcolor{green!25}4.2  \\
carton\_milk       & 24.9 & 27.5  & 29.8 & \cellcolor{green!25}38.3 & 0.0  & \cellcolor{green!25}35.3 & 17.6 & \cellcolor{yellow!25}29.4 & 0.0  & 15.8  & 6.0   & \cellcolor{yellow!25}14.5 \\
carton\_oj         & 5.5  & \cellcolor{yellow!25}35.0 & 25.6 & \cellcolor{green!25}39.3 & 0.0  & 31.2  & \cellcolor{yellow!25}43.8 & \cellcolor{green!25}43.8 & 0.0  & \cellcolor{yellow!25}16.2 & 14.4  & \cellcolor{green!25}20.6 \\
cellphone          & \cellcolor{green!25}49.3 & 8.3  & 17.6 & \cellcolor{yellow!25}23.0 & \cellcolor{green!25}42.3 & 9.6  & 23.1 & \cellcolor{yellow!25}25.0 & \cellcolor{green!25}31.6 & 6.7   & 8.5   & \cellcolor{yellow!25}14.0 \\
coffee\_pot        & 27.1 & \cellcolor{yellow!25}42.4 & 35.5 & \cellcolor{green!25}43.7 & 18.4 & 57.1  & \cellcolor{yellow!25}71.4 & \cellcolor{green!25}75.5 & 15.3 & \cellcolor{yellow!25}29.1 & 28.1  & \cellcolor{green!25}35.2 \\
dino\_toy          & 24.8 & 15.9  & \cellcolor{yellow!25}38.6 & \cellcolor{green!25}40.5 & 33.3 & 66.7  & \cellcolor{yellow!25}72.2 & \cellcolor{green!25}77.8 & 24.8 & \cellcolor{yellow!25}32.7 & \cellcolor{yellow!25}34.5 & \cellcolor{green!25}39.7 \\
food\_vegetables   & \cellcolor{yellow!25}43.0 & 41.6 & 40.7 & \cellcolor{green!25}48.9 & 9.5  & \cellcolor{yellow!25}23.8 & \cellcolor{green!25}33.3 & 23.8  & 8.4  & \cellcolor{yellow!25}14.1 & \cellcolor{green!25}18.7 & 13.2 \\
keyboard           & \cellcolor{green!25}37.6 & \cellcolor{yellow!25}31.8 & 21.9 & 31.4 & 60.9 & 84.1  & \cellcolor{green!25}92.8 & \cellcolor{yellow!25}91.3 & \cellcolor{green!25}35.6 & \cellcolor{yellow!25}29.4 & 19.8  & \cellcolor{green!25}28.1 \\
mouse              & 48.3 & \cellcolor{green!25}59.3 & 44.5 & \cellcolor{yellow!25}56.3 & 18.6 & 39.5  & \cellcolor{yellow!25}67.4 & \cellcolor{green!25}72.1 & 13.8 & 28.9  & \cellcolor{yellow!25}34.1 & \cellcolor{green!25}43.8 \\
mug\_white         & 15.2 & \cellcolor{yellow!25}33.1 & 30.9 & \cellcolor{green!25}42.4 & 11.1 & 28.9  & \cellcolor{yellow!25}37.8 & \cellcolor{green!25}57.8 & 8.1  & \cellcolor{yellow!25}15.9 & \cellcolor{yellow!25}17.5 & \cellcolor{green!25}28.9 \\
plate\_bamboo      & 36.6 & \cellcolor{green!25}56.3 & 33.6 & \cellcolor{yellow!25}55.1 & 35.6 & 81.4  & \cellcolor{yellow!25}89.8 & \cellcolor{green!25}93.2 & 24.4 & \cellcolor{yellow!25}50.0 & 30.5 & \cellcolor{green!25}50.5 \\
potato\_masher     & 2.2  & 5.3   & \cellcolor{yellow!25}17.7 & \cellcolor{green!25}26.5 & 4.9  & \cellcolor{yellow!25}50.5 & 47.6 & \cellcolor{yellow!25}29.4 & 2.9  & \cellcolor{green!25}28.3 & 15.4 & \cellcolor{yellow!25}25.2 \\
puzzle\_toy        & 37.1 & \cellcolor{green!25}46.4 & 32.1 & \cellcolor{yellow!25}44.8 & 24.4 & \cellcolor{yellow!25}54.9 & 51.2 & \cellcolor{green!25}64.6 & 17.9 & \cellcolor{yellow!25}29.2 & 19.1 & \cellcolor{green!25}31.7 \\
spatula\_red       & 1.3  & 14.6  & \cellcolor{yellow!25}26.7 & \cellcolor{green!25}35.4 & 4.8  & 50.8  & \cellcolor{yellow!25}76.2 & \cellcolor{yellow!25}85.7 & 3.3  & \cellcolor{yellow!25}33.0 & 24.6 & \cellcolor{green!25}35.4 \\
whiteboard\_eraser & 23.0 & \cellcolor{green!25}42.6 & 36.0 & \cellcolor{yellow!25}42.1 & 20.0 & 80.0  & \cellcolor{yellow!25}100.0 & \cellcolor{green!25}100.0 & 9.9  & \cellcolor{yellow!25}37.8 & 36.0 & \cellcolor{green!25}42.1 \\
whiteboard\_marker  & 30.4 & \cellcolor{green!25}49.2 & 37.4 &  & 0.0  & \cellcolor{green!25}100.0 & 66.7 & \cellcolor{yellow!25}45.7 & 0.0  & \cellcolor{yellow!25}71.2 & 27.3 & \cellcolor{yellow!25}35.4 \\
\midrule
\textit{Average}   & 27.8 & \cellcolor{yellow!25}37.4 & 30.8 & \cellcolor{green!25}\textbf{42.0} & 16.8 & 43.0  & \cellcolor{yellow!25}51.9 & \cellcolor{green!25}\textbf{58.1} & 11.5 & \cellcolor{yellow!25}22.9 & 18.4 & \cellcolor{green!25}\textbf{26.8} \\
\bottomrule
\end{tabular}

    }
    \label{tab:hot3d_stable_grasps_results}
    \caption{\textbf{Results on \StableGrasp in EPIC-HIT.}
    \colorbox{green!25}{Green} for \colorbox{green!25}{best} and \colorbox{yellow!25}{yellow} shows the \colorbox{yellow!25}{second} best.
    $\text{COP}^{\dagger}$ is COP w/o propagation.
    }
    \vspace*{-12pt}
    \centering
    \resizebox{\textwidth}{!}{%
    \setlength{\tabcolsep}{10pt} %

    \begin{tabular}{@{}lcccccccccccc}
    \toprule
    \multirow{2}{*}{Category} & \multicolumn{4}{c}{IOU} & \multicolumn{4}{c}{SCA@0.8} & \multicolumn{4}{c}{SCA@0.6} \\
    \cmidrule(lr){2-5}\cmidrule(lr){6-9}\cmidrule(lr){10-13}
     & HOMan~\cite{Hasson2021TowardsVideos} & Rigid~\cite{swamy2023showme} & Dynamic & $\text{COP}^{\dagger}$ & HOMan~\cite{Hasson2021TowardsVideos} & Rigid~\cite{swamy2023showme} & Dynamic & $\text{COP}^{\dagger}$ & HOMan~\cite{Hasson2021TowardsVideos} & Rigid~\cite{swamy2023showme} & Dynamic & $\text{COP}^{\dagger}$ \\
    \midrule
    bottle   & 56.6 & 66.0 & \cellcolor{green!25}75.5 & \cellcolor{yellow!25}72.2 & 3.5  & 16.4 & \cellcolor{yellow!25}21.8 & \cellcolor{green!25}29.6 & 5.7  & \cellcolor{yellow!25}51.9 & 36.6 & \cellcolor{green!25}56.3 \\
    bowl     & 54.2 & 52.9 & \cellcolor{green!25}57.9 & \cellcolor{yellow!25}55.1 & 1.0  & \cellcolor{yellow!25}10.5 & 9.3  & \cellcolor{green!25}11.9 & 1.6  & \cellcolor{yellow!25}28.2 & 19.7 & \cellcolor{green!25}32.0 \\
    can      & 47.5 & 49.8 & \cellcolor{green!25}55.2 & \cellcolor{yellow!25}53.6 & 3.5  & 12.7 & \cellcolor{yellow!25}13.0 & \cellcolor{green!25}16.5 & 5.3  & \cellcolor{yellow!25}24.1 & 19.4 & \cellcolor{green!25}28.1 \\
    cup      & 56.9 & 61.4 & \cellcolor{green!25}67.5 & \cellcolor{yellow!25}66.1 & 5.5  & 12.4 & \cellcolor{yellow!25}16.5 & \cellcolor{green!25}20.0 & 7.8  & \cellcolor{yellow!25}43.2 & 38.9 & \cellcolor{green!25}49.7 \\
    glass    & 55.4 & 57.6 & \cellcolor{green!25}65.2 & \cellcolor{yellow!25}63.1 & 3.0  & 10.0 & \cellcolor{yellow!25}14.8 & \cellcolor{green!25}15.9 & 4.2  & \cellcolor{yellow!25}36.3 & 30.1 & \cellcolor{green!25}40.9 \\
    mug      & 59.4 & 58.1 & \cellcolor{green!25}63.8 & \cellcolor{yellow!25}62.1 & 3.9  & \cellcolor{yellow!25}6.5  & 6.3  & \cellcolor{green!25}9.2  & 6.6  & \cellcolor{yellow!25}36.3 & 29.6 & \cellcolor{green!25}42.5 \\
    pan      & \cellcolor{yellow!25}48.3 & 45.8 & \cellcolor{green!25}48.9 & 42.4 & 0.5  & \cellcolor{green!25}4.5  & \cellcolor{yellow!25}4.3  & 3.5  & 1.9  & \cellcolor{green!25}18.6 & 12.8 & \cellcolor{yellow!25}15.1 \\
    plate    & 61.1 & 61.5 & \cellcolor{green!25}68.1 & \cellcolor{yellow!25}65.3 & 1.0  & 15.5 & \cellcolor{yellow!25}17.6 & \cellcolor{green!25}22.5 & 1.6  & \cellcolor{yellow!25}39.3 & 30.1 & \cellcolor{green!25}46.4 \\
    saucepan & 51.1 & 53.1 & \cellcolor{green!25}57.5 & \cellcolor{yellow!25}54.4 & 0.4  & 3.1  & \cellcolor{green!25}5.4  & \cellcolor{yellow!25}5.3  & 2.4  & \cellcolor{green!25}32.7 & 25.7 & \cellcolor{yellow!25}31.1 \\
    \midrule
    \textit{Average} & 54.8 & 55.8 & \cellcolor{green!25}\textbf{61.7} & \cellcolor{yellow!25}58.2 & 1.9  & 10.6 & \cellcolor{yellow!25}12.2 & \cellcolor{green!25}\textbf{15.0} & 3.3  & \cellcolor{yellow!25}33.4 & 25.2 & \cellcolor{green!25}\textbf{36.5} \\
    \bottomrule
    \end{tabular}

     }
     \vspace*{-12pt}
    \label{tab:epic_sg_results}
\end{table*}

\noindent \textbf{Average Stable Contact Area at ADD Success (SCA-ADD)}. 
When a pose is considered correct for a sequence, \ie average distance is within 10\% of the object's diameter and thus ADD is 1, we measure the stable contact area across the sequence, defined as the average IOU of in contact area between all pairs of frames~(\cref{sec:hoi_timeline_definition}). SCA-ADD is set to 0 when ADD is 0 (average distance below threshold). We average SCA-ADD over all examples. We use this metric to showcase our ability to reconstruct stable grasps.

\noindent \textbf{Average Stable Contact Area at high IOU (SCA-IOU)}. 
We analogously report SCA when IOU is more than certain thresholds. 
We use 80\% as threshold on HOT3D, and report both 80\% and 60\% on EPIC. 
SCA-IOU is set to 0 when IOU is below the threshold. We report average SCA-IOU.

\noindent \textbf{IOU vs. SCA-IOU}. Note that IOU and SCA-IOU can be contradictory. A method can maximise IOU by individually fitting to each mask, resulting in a lower SCA-IOU.
Using both metrics allows us to understand the performance of different baselines versus our proposed \meth.

\begin{figure}
    \centering
    \includegraphics[width=\linewidth]{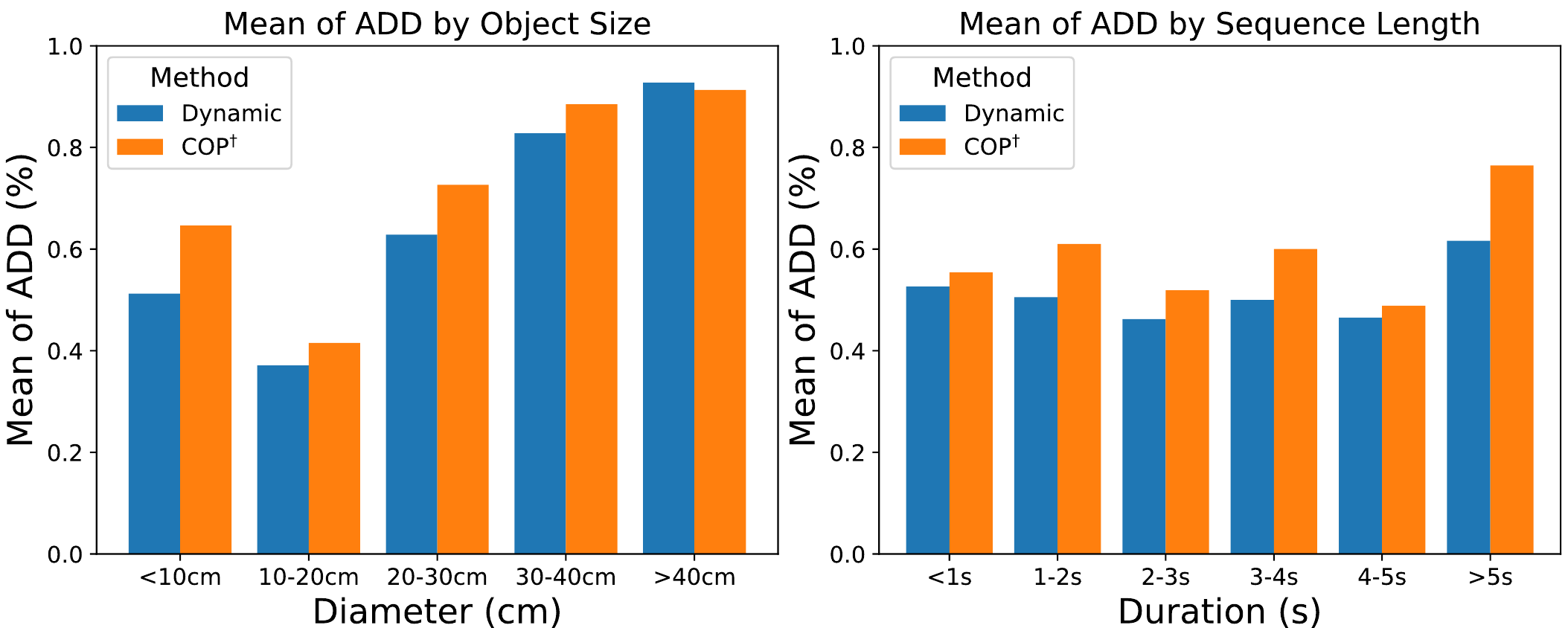}
    \vspace*{-16pt}
    \caption{Improvement of COP over Dynamic method for different object sizes (left) and \StableGrasp lengths (right).}
    \vspace*{-16pt}
\label{fig:analysis_SGonly}
\end{figure}

\subsection{Results and Ablation}
\label{sec:qualitative_results}

\noindent\textbf{\StableGrasp in HOT3D-HIT.}
We first study the reconstruction of \StableGrasp segments. To ensure fair comparison, here \meth does not employ propagation.
\Cref{tab:hot3d_stable_grasps_results} contains per-category results for \StableGrasp in HOT3D-HIT.
On average, COP improves ADD from $51.9\%$ with the dynamic assumption to $58.1\%$ within the \StableGrasp,
and improves SCA-ADD from $22.9\%$ with the rigid assumption to $26.8\%$.
Categories like ``potato\_masher'' significantly improve in ADD score ($+16.5\%$).
The baseline `Rigid' has high SCA-ADD, but ADD is significantly lower than \meth.
HOMan~\cite{Hasson2021TowardsVideos} performs badly on most categories as the contact is not maintained through its iterative optimisation.

In \cref{fig:analysis_SGonly}, we investigate the impact of \StableGrasp sequence length and different object sizes.
\meth consistently outperforms Dynamic, only dropping slightly for large objects in HOT3D.
Moreover, \meth performs significantly better for smaller objects and on long sequences.

\begin{figure}[!t]
\centering
\includegraphics[width=1\linewidth]{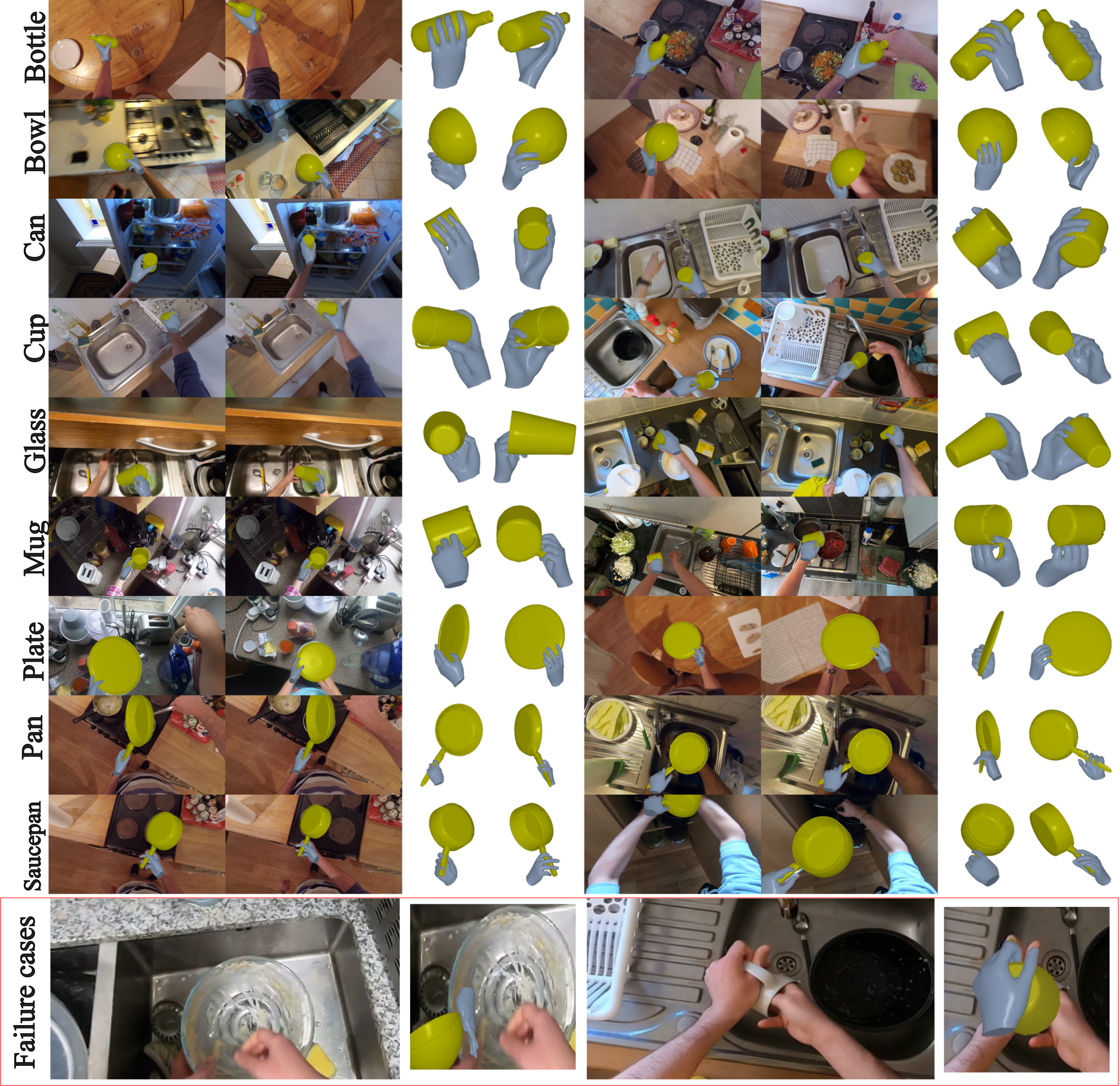}
\vspace*{-18pt}
\caption{\textbf{Qualitative Results on \StableGrasp in EPIC-\timeline} (2 examples/category): projected reconstruction results and reconstruction in rotated views.
Bottom: failure cases due to wrong hand pose (left) and extreme occlusion (right).
}
\vspace*{-12pt}
    \label{fig:bigqual}
\end{figure}

\noindent\textbf{\StableGrasp in EPIC-HIT.}
\Cref{tab:epic_sg_results} contains analogous per-category results for \StableGrasp in EPIC-HIT.
\meth outperforms baselines on the SCA metric. %
Dynamic performs well on IOU, as it fits to each mask individually but \textit{does not} maintain a stable grasp indicated by the lower SCA metric. 
Note that IOU is the 2D proxy metric of pose accuracy;
$E_{SG}$ regularises the IOU fitting to improve SCA. 
Rigid achieves best SCA results on the ``pan'' category only -- indicating the functional hold of the pan does not allow finger articulations.
\Cref{fig:bigqual} shows reconstruction results of \meth on \StableGrasp in EPIC-HIT.

\noindent \textbf{\timeline in HOT3D-HIT.}
In~\cref{tab:results_hot3d_timeline}, 
we show quantitative results on the full timelines (HIT) reconstruction on HOT3D-HIT.
While propagation could theoretically improve the Dynamic baseline, we focus our propagation ablation on COP.
Improvements can be seen across the board -- ADD in \StableGrasp  segments is improved by $12.2\%$ and \Static segments benefits also, improving by $14.5\%$.
This highlights that \meth with propagation improves the reconstruction across all segment types.

\begin{figure}
    \centering\includegraphics[width=\linewidth]{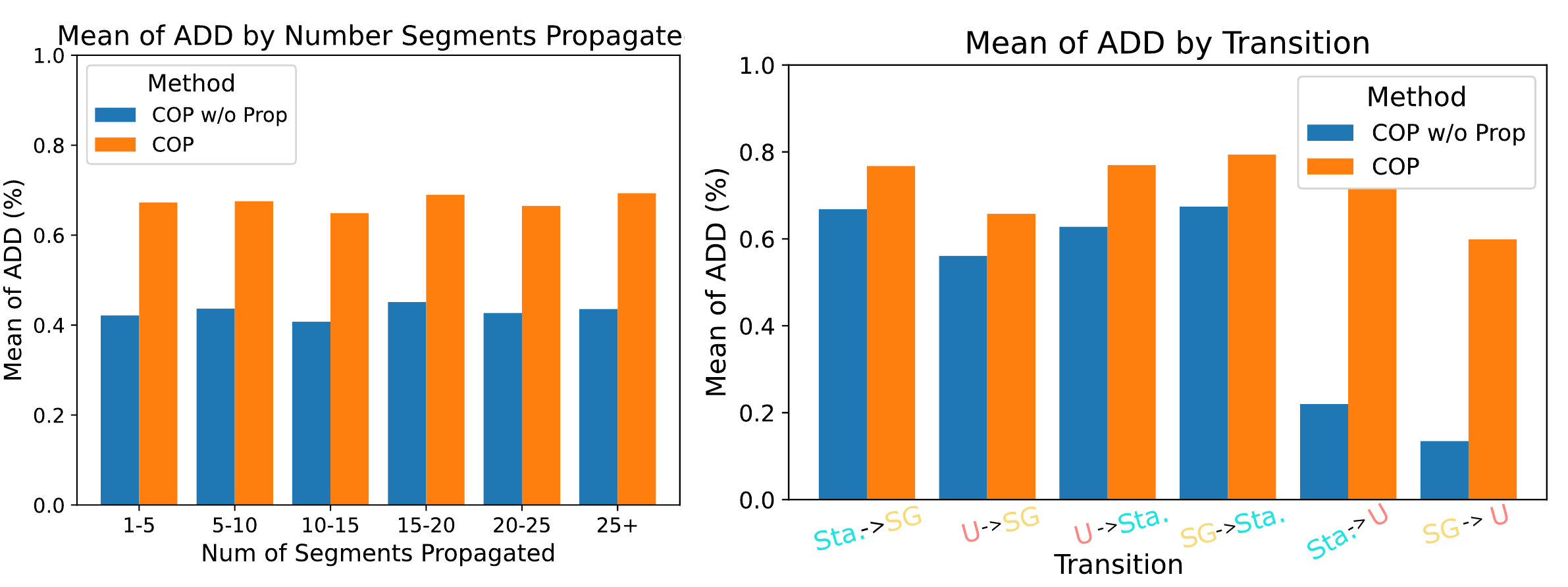}
    \vspace*{-16pt}
    \caption{Impact of propagation on segment count and types.}
    \vspace*{-12pt}
\label{fig:analysis_prop_results}
\end{figure}

\Cref{fig:analysis_prop_results} summarises the impact of propagation on HITs with different segment count and types.
Improvement is consistent across counts.
Notably, transition to \UnstableContact shows significant improvement in ADD.

\noindent \textbf{\timeline in EPIC-HIT.} In~\cref{tab:results_epic_timeline} we report analogous results on  EPIC-HIT.
\meth again shows consistent improvements over all metrics across different types of segments.
We show qualitative results for \timeline reconstruction in~\cref{fig:qual_teaser}.

\noindent \textbf{Ablation on the constraint in~\cref{eq:e_SG}}.  
Recall that in $E_{SG}$, 
we first index every object vertex $v_o \in V_o$, compute its distance to each finger tips vertex, then constrain the variation over all pairs of frames. 
In~\cref{tab:ablate_E_SG_variants_new}, we ablate two variants, aiming to maximise the stable contact area:
i) select the nearest object vertex $v^*_o$ corresponds to each $v_h \in V_F$, instead of selecting all object vertices $V_o$;
ii) select $N$ consecutive frames to constrain $d_{oh}$.
The results show that our constraint performs the best.

\begin{table}[t]
    \centering
\caption{Results on HOT3D-HIT}
\vspace*{-10pt}
\resizebox{\linewidth}{!}{
    \begin{tabular}{@{}l cc c c@{}}
    \toprule
    \multirow{2}{*}{\textbf{Method}} & 
    \multicolumn{2}{c}{\textbf{\StableGrasp}} & 
    {\textbf{\Static}} & 
    {\textbf{\UnstableContact}} \\
    \cmidrule(lr){2-3} \cmidrule(lr){4-4} \cmidrule(l){5-5}
      & \textbf{ADD} & \textbf{SCA-ADD} &  \textbf{ADD} & \textbf{ADD} \\
    \midrule
    COP \textit{w/o Propagation}   & 58.1 &   26.8  & 64.7 & 17.3 \\
    COP                            & \textbf{70.3} &  \textbf{34.1}  & \textbf{79.2}  & \textbf{67.1} \\
    \bottomrule
    \end{tabular}
    }
\label{tab:results_hot3d_timeline}
\caption{Results on EPIC-HIT}
\vspace*{-10pt}
\resizebox{\linewidth}{!}{
    \begin{tabular}{@{}l cc c c@{}}
    \toprule
    \multirow{2}{*}{\textbf{Method}} & 
    \multicolumn{2}{c}{\textbf{\StableGrasp}} & 
    {\textbf{\Static}} & 
    {\textbf{\UnstableContact}} \\
    \cmidrule(lr){2-3} \cmidrule(lr){4-4} \cmidrule(l){5-5}
      & \textbf{IOU} & \textbf{SCA@0.8} &  \textbf{IOU} & \textbf{IOU} \\
    \midrule
    COP \textit{w/o Propagation}   & 68.0 &   23.8  & 65.8 & 74.3 \\
    COP                  & \textbf{69.2} &  \textbf{25.5}  & \textbf{68.3}  & \textbf{79.0} \\
    \bottomrule
    \end{tabular}
    }
    \vspace*{-6pt}
\label{tab:results_epic_timeline}
\end{table}

\begin{table}[t]
    \caption{Ablation on the \StableGrasp Loss $E_{SG}$ variants on HOT3D. We show improvement over the Dynamic Baseline}
    \centering
\resizebox{\linewidth}{!}{
    \begin{tabular}{cccc}
    \toprule
    {Obj. Vert. Selection}& {Frames Selection} & ADD & SCA-ADD \\
    \midrule
    \rowcolor{blue!10} $V_o$ & $N^2$ pairs      & 58.1 (+6.2) & 26.8  \\
    $v^*_o$ & $N^2$ pairs                       & 52.9 (+1.0) & 21.3  \\
    $V_o$ & $N$ consecutive                     & 54.2 (+2.3) & 22.5  \\
    \hline
    \multicolumn{2}{c}{Dynamic Baseline}        & 51.9 (+0.0) & 18.4  \\
    \bottomrule
    \end{tabular}    
    }
    \vspace*{-6pt}\label{tab:ablate_E_SG_variants_new}
\end{table}
\noindent \textbf{Ablation on the weights in~\cref{eq:totError}}. In~\cref{tab:ablate_weights_new}, we ablate the weights $\lambda_1$ and $\lambda_2$ introduced in the loss function~\cref{eq:totError}.
The results suggest that $E_{SG}$ is important while the physical heuristics $E_{pull}$ and $E_{push}$ are also necessary.
Dynamic is equivalent to $\lambda_1 = 0$ and $\lambda_2 = 0.1$.

\noindent{\textbf{Additional Results.}} In supp., we (i) ablate the robustness to noise in the segment boundaries along the \timeline; (ii) report results on stable grasp segments from the ARCTIC dataset~\cite{fan2023arctic} with matching conclusions; and (iii) show further examples of failure of CAD-Agnostic methods on in-the-wild recordings. 

\begin{table}[t]
    \caption{\textbf{Ablation on the loss weights.}
    Chosen $\lambda_1$ and $\lambda_2$ in blue}
    \centering
\resizebox{0.5\linewidth}{!}{
    \begin{tabular}{cccccc}
    \toprule
    {$\lambda_1$}& {$\lambda_2$} & ADD & SCA-ADD \\
    \midrule
        0 & 0.1 & 51.9 & 18.4  \\
        1 & 0 & 54.8 & 25.0  \\
        \rowcolor{blue!10} 1 & 0.1 & 58.1 & 26.8 \\
        10 & 0.1 & 51.5 & 26.6 \\
    \bottomrule
    \end{tabular}
    }
    \vspace*{-6pt}
    \label{tab:ablate_weights_new}
\end{table}

\section{Conclusion}

We proposed the \taskfull (\task) task -- which aims to reconstruct an object along time including when the object is static in the scene, in an stable or unstable grasp. 
To tackle the task, we propose the \methfull (\meth) framework which builds around \StableGrasp and propagates poses across segments for superior reconstruction.
We propose HOT3D-\timeline (with 3D ground truth) and EPIC-\timeline (in-the-wild) datasets to evaluate \meth.
We highlight the efficacy of stable grasp and pose propagation on both datasets.
By reconstructing full timelines, we hope to encourage future works to quantitatively evaluate timeline reconstruction methods in-the-wild.

\noindent \textbf{Acknowledgements}
This work is supported by EPSRC UMPIRE EP/T004991/1.
Z Zhu is supported by UoB-CSC scholarship.
S Bansal is supported by a Charitable
Donation to the University of Bristol from Meta.
S. Tripathi is supported by the International Max Planck Research School for Intelligent Systems (IMPRS-IS).
We thank Ahmad Darkhalil for help with VISOR masks.

{
    \small
    \bibliographystyle{ieeenat_fullname}
    \bibliography{zhifan_rebibed}

@String(IJCV   = {International Journal of Computer Vision})

@String(CVPR   = {Proceedings of the IEEE Conference on Computer Vision and Pattern Recognition})

@String(ICCV   = {Proceedings of the IEEE International Conference on Computer Vision})

@String(ICCVW   = {Proceedings of the IEEE International Conference on Computer Vision Workshops})

@String(ECCV   = {Proceedings of the European Conference on Computer Vision})

@String(ECCVW   = {Proceedings of the European Conference on Computer Vision Workshops})

@String(AAAI   = {Proceedings of the AAAI Conference on Artificial Intelligence})

@String(RSS = {Proceedings of Robotics: Science and Systems})

@String(NeurIPS = {Advances in Neural Information Processing Systems})

@inproceedings{Chen2022AlignSDF:Reconstruction,
  title={AlignSDF: Pose-Aligned Signed Distance Fields for Hand-Object Reconstruction},
  author={Chen, Zerui and Hasson, Yana and Schmid, Cordelia and Laptev, Ivan},
  booktitle = ECCV,
  pages={231--248},
  year={2022},
}

@inproceedings{Yang2022Artiboost,
  title={{ArtiBoost: Boosting articulated 3d hand-object pose estimation via online exploration and synthesis}},
  author={Yang, Lixin and Li, Kailin and Zhan, Xinyu and Lv, Jun and Xu, Wenqiang and Li, Jiefeng and Lu, Cewu},
  booktitle=CVPR,
  pages={2750--2760},
  year={2022}
}

@inproceedings{SenerAssembly101:Activities,
  title={Assembly101: A large-scale multi-view video dataset for understanding procedural activities},
  author={Sener, Fadime and Chatterjee, Dibyadip and Shelepov, Daniel and He, Kun and Singhania, Dipika and Wang, Robert and Yao, Angela},
  booktitle={Proceedings of the IEEE/CVF Conference on Computer Vision and Pattern Recognition},
  pages={21096--21106},
  year={2022}
}

@inproceedings{BrahmbhattContactPose:Pose,
  title={ContactPose: A dataset of grasps with object contact and hand pose},
  author={Brahmbhatt, Samarth and Tang, Chengcheng and Twigg, Christopher D and Kemp, Charles C and Hays, James},
  booktitle=ECCV,
  pages={361--378},
  year={2020},
}

@inproceedings{LomonacoCORe50:Recognition,
  title={Core50: a new dataset and benchmark for continuous object recognition},
  author={Lomonaco, Vincenzo and Maltoni, Davide},
  booktitle={Conference on Robot Learning},
  pages={17--26},
  year={2017},
}

@inproceedings{Yang2020CPF:Interaction,
  title={{CPF: Learning a contact potential field to model the hand-object interaction}},
  author={Yang, Lixin and Zhan, Xinyu and Li, Kailin and Xu, Wenqiang and Li, Jiefeng and Lu, Cewu},
  booktitle=ICCV,
  pages={11097--11106},
  year={2021}
}

@inproceedings{ChaoDexYCB:Objects,
  title={DexYCB: A benchmark for capturing hand grasping of objects},
  author={Chao, Yu-Wei and Yang, Wei and Xiang, Yu and Molchanov, Pavlo and Handa, Ankur and Tremblay, Jonathan and Narang, Yashraj S and Van Wyk, Karl and Iqbal, Umar and Birchfield, Stan and others},
  booktitle=CVPR,
  pages={9044--9053},
  year={2021}
}

@article{Romero2017EmbodiedTogether,
 author = {Romero, Javier and Tzionas, Dimitrios and Black, Michael J},
 journal = {ACM Trans. Graph},
 pages = {17},
 title = {{Embodied Hands: Modeling and Capturing Hands and Bodies Together}},
 volume = {36},
 year = {2017}
}

@inproceedings{VISOR2022,
   title={EPIC-KITCHENS VISOR Benchmark: VIdeo Segmentations and Object Relations},
   author={Darkhalil, Ahmad and Shan, Dandan and Zhu, Bin and Ma, Jian and Kar, Amlan and Higgins, Richard and Fidler, Sanja and Fouhey, David and Damen, Dima},
   booktitle   = NeurIPS,
   year      = {2022}
}

@inproceedings{Garcia-HernandoFirst-PersonAnnotations,
 author = {Guillermo Garcia{-}Hernando and
Shanxin Yuan and
Seungryul Baek and
Tae{-}Kyun Kim},
 bibsource = {dblp computer science bibliography, https://dblp.org},
 booktitle = {Proceedings of the {IEEE} Conference on Computer Vision and Pattern Recognition},
 pages = {409--419},
 title = {First-Person Hand Action Benchmark With {RGB-D} Videos and 3D Hand
Pose Annotations},
 year = {2018}
}

@inproceedings{RongFrankMocap:Integration,
  title={Frankmocap: A monocular 3d whole-body pose estimation system via regression and integration},
  author={Rong, Yu and Shiratori, Takaaki and Joo, Hanbyul},
  booktitle=ICCVW,
  pages={1749--1759},
  year={2021}
}

@inproceedings{TaheriGRAB:Objects,
  title={GRAB: A dataset of whole-body human grasping of objects},
  author={Taheri, Omid and Ghorbani, Nima and Black, Michael J and Tzionas, Dimitrios},
  booktitle= ECCV,
  pages={581--600},
  year={2020},
}

@inproceedings{Karunratanakul2020GraspingGrasps,
  title={Grasping field: Learning implicit representations for human grasps},
  author={Karunratanakul, Korrawe and Yang, Jinlong and Zhang, Yan and Black, Michael J and Muandet, Krikamol and Tang, Siyu},
  booktitle={International Conference on 3D Vision (3DV)},
  pages={333--344},
  year={2020},
}

@inproceedings{kwon2021h2o,
  title={H2o: Two hands manipulating objects for first person interaction recognition},
  author={Kwon, Taein and Tekin, Bugra and St{\"u}hmer, Jan and Bogo, Federica and Pollefeys, Marc},
  booktitle=ICCV,
  pages={10138--10148},
  year={2021}
}

@inproceedings{Liu2022HOI4D:Interaction,
  title={{HOI4D: A 4D Egocentric Dataset for Category-Level Human-Object Interaction}},
  author={Liu, Yunze and Liu, Yun and Jiang, Che and Lyu, Kangbo and Wan, Weikang and Shen, Hao and Liang, Boqiang and Fu, Zhoujie and Wang, He and Yi, Li},
  booktitle=CVPR,
  pages={21013--21022},
  year={2022}
}

@inproceedings{HampaliHOnnotate:Poses,
 author = {Shreyas Hampali and
Mahdi Rad and
Markus Oberweger and
Vincent Lepetit},
 bibsource = {dblp computer science bibliography, https://dblp.org},
 booktitle = CVPR,
 pages = {3193--3203},
 title = {HOnnotate: {A} Method for 3D Annotation of Hand and Object Poses},
 year = {2020}
}

@inproceedings{HassonLearningObjects,
 author = {Yana Hasson and
G{\"{u}}l Varol and
Dimitrios Tzionas and
Igor Kalevatykh and
Michael J. Black and
Ivan Laptev and
Cordelia Schmid},
 bibsource = {dblp computer science bibliography, https://dblp.org},
 booktitle = {Proceedings of the {IEEE} Conference on Computer Vision and Pattern Recognition},
 pages = {11807--11816},
 title = {Learning Joint Reconstruction of Hands and Manipulated Objects},
 year = {2019}
}

@article{Patel2022LearningVideos,
  title={Learning to Imitate Object Interactions from Internet Videos},
  author={Patel, Austin and Wang, Andrew and Radosavovic, Ilija and Malik, Jitendra},
  journal={arXiv preprint arXiv:2211.13225},
  year={2022}
}

@article{Zhang2021ManipNet:Representation,
 author = {Zhang, He and Ye, Yuting and Shiratori, Takaaki and Komura, Taku},
 issn = {15577368},
 journal = {ACM Transactions on Graphics},
 number = {4},
 publisher = {Association for Computing Machinery},
 title = {{ManipNet: Neural Manipulation Synthesis with a Hand-Object Spatial Representation}},
 volume = {40},
 year = {2021}
}

@inproceedings{KatoNeuralRenderer,
 author = {Hiroharu Kato and
Yoshitaka Ushiku and
Tatsuya Harada},
 bibsource = {dblp computer science bibliography, https://dblp.org},
 booktitle = {Proceedings of the {IEEE} Conference on Computer Vision and Pattern Recognition},
 pages = {3907--3916},
 title = {Neural 3D Mesh Renderer},
 year = {2018}
}

@inproceedings{Sucar2020NodeSLAM:Reconstruction,
 arxivid = {2004.04485v2},
 author = {Sucar, Edgar and Wada, Kentaro and Davison, Andrew},
 booktitle = {Proceedings of the International Conference on 3D Vision (3DV)},
 title = {{NodeSLAM: Neural Object Descriptors for Multi-View Shape Reconstruction}},
 year = {2020}
}

@inproceedings{YangOakInk:Interaction,
  title={OakInk: A Large-scale Knowledge Repository for Understanding Hand-Object Interaction},
  author={Yang, Lixin and Li, Kailin and Zhan, Xinyu and Wu, Fei and Xu, Anran and Liu, Liu and Lu, Cewu},
  booktitle={Proceedings of the IEEE/CVF Conference on Computer Vision and Pattern Recognition},
  pages={20953--20962},
  year={2022}
}

@inproceedings{ZhangPerceivingWild,
  title={Perceiving 3d human-object spatial arrangements from a single image in the wild},
  author={Zhang, Jason Y and Pepose, Sam and Joo, Hanbyul and Ramanan, Deva and Malik, Jitendra and Kanazawa, Angjoo},
  booktitle=ECCV,
  pages={34--51},
  year={2020},
}

@inproceedings{Huang2022ReconstructingVideo,
 arxivid = {2211.16835},
 author = {Huang, Di and Ji, Xiaopeng and He, Xingyi and Sun, Jiaming and He, Tong and Shuai, Qing and Ouyang, Wanli and Zhou, Xiaowei},
 booktitle = {Proceedings of SIGGRAPH Asia 2022 Conference Papers},
 isbn = {9781450394703},
 title = {{Reconstructing Hand-Held Objects from Monocular Video}},
 year = {2022}
}

@inproceedings{CaoReconstructingWild,
  title={Reconstructing hand-object interactions in the wild},
  author={Cao, Zhe and Radosavovic, Ilija and Kanazawa, Angjoo and Malik, Jitendra},
  booktitle= ICCV,
  pages={12417--12426},
  year={2021}
}

@ARTICLE{Damen2022RESCALING,
   title={Rescaling Egocentric Vision: Collection, Pipeline and Challenges for EPIC-KITCHENS-100},
   author={Damen, Dima and Doughty, Hazel and Maria Farinella, Giovanni  and and Furnari, Antonino 
   and Ma, Jian and Kazakos, Evangelos and Moltisanti, Davide and Munro, Jonathan 
   and Perrett, Toby and Price, Will and Wray, Michael},
   journal   = {International Journal of Computer Vision (IJCV)},
   year      = {2022},
   volume = {130},
   pages = {33–55},
}

@inproceedings{DamenScalingDataset,
  title={Scaling egocentric vision: The epic-kitchens dataset},
  author={Damen, Dima and Doughty, Hazel and Farinella, Giovanni Maria and Fidler, Sanja and Furnari, Antonino and Kazakos, Evangelos and Moltisanti, Davide and Munro, Jonathan and Perrett, Toby and Price, Will and others},
  booktitle=ECCV,
  pages={720--736},
  year={2018}
}

@inproceedings{LiuSemi-SupervisedTime,
  title={Semi-supervised 3d hand-object poses estimation with interactions in time},
  author={Liu, Shaowei and Jiang, Hanwen and Xu, Jiarui and Liu, Sifei and Wang, Xiaolong},
  booktitle={Proceedings of the IEEE/CVF Conference on Computer Vision and Pattern Recognition},
  pages={14687--14697},
  year={2021}
}

@inproceedings{SchonbergerStructure-from-MotionRevisited,
 author = {Johannes L. Sch{\"{o}}nberger and
Jan{-}Michael Frahm},
 bibsource = {dblp computer science bibliography, https://dblp.org},
 booktitle = CVPR,
 doi = {10.1109/CVPR.2016.445},
 pages = {4104--4113},
 publisher = {{IEEE} Computer Society},
 title = {Structure-from-Motion Revisited},
 url = {https://doi.org/10.1109/CVPR.2016.445},
 year = {2016}
}

@article{Feix2016TheTypes,
 author = {Feix, Thomas and Romero, Javier and Schmiedmayer, Heinz Bodo and Dollar, Aaron M. and Kragic, Danica},
 issn = {21682291},
 journal = {IEEE Transactions on Human-Machine Systems},
 number = {1},
 pages = {66--77},
 publisher = {Institute of Electrical and Electronics Engineers Inc.},
 title = {{The GRASP Taxonomy of Human Grasp Types}},
 volume = {46},
 year = {2016}
}

@inproceedings{Hasson2021TowardsVideos,
  title={Towards unconstrained joint hand-object reconstruction from RGB videos},
  author={Hasson, Yana and Varol, G{\"u}l and Schmid, Cordelia and Laptev, Ivan},
  booktitle={International Conference on 3D Vision (3DV)},
  pages={659--668},
  year={2021},
}

@inproceedings{YeWhatsHands,
  title={What's in your hands? 3D Reconstruction of Generic Objects in Hands},
  author={Ye, Yufei and Gupta, Abhinav and Tulsiani, Shubham},
  booktitle=CVPR,
  pages={3895--3905},
  year={2022}
}

@inproceedings{Tse2022Collaborative,
  title={Collaborative learning for hand and object reconstruction with attention-guided graph convolution},
  author={Tse, Tze Ho Elden and Kim, Kwang In and Leonardis, Ales and Chang, Hyung Jin},
  booktitle=CVPR,
  pages={1664--1674},
  year={2022}
}

@inproceedings{Wang2023Interacting,
  title={Interacting Hand-Object Pose Estimation via Dense Mutual Attention},
  author={Wang, Rong and Mao, Wei and Li, Hongdong},
  booktitle={Proceedings of the IEEE/CVF Winter Conference on Applications of Computer Vision},
  pages={5735--5745},
  year={2023}
}

@inproceedings{Aboukhadra2023Thor,
  title={THOR-Net: End-to-end Graformer-based Realistic Two Hands and Object Reconstruction with Self-supervision},
  author={Aboukhadra, Ahmed Tawfik and Malik, Jameel and Elhayek, Ahmed and Robertini, Nadia and Stricker, Didier},
  booktitle={Proceedings of the IEEE/CVF Winter Conference on Applications of Computer Vision},
  pages={1001--1010},
  year={2023}
}

@inproceedings{fan2023arctic,
  title = {{ARCTIC}: A Dataset for Dexterous Bimanual Hand-Object Manipulation},
  author = {Fan, Zicong and Taheri, Omid and Tzionas, Dimitrios and Kocabas, Muhammed and Kaufmann, Manuel and Black, Michael J. and Hilliges, Otmar},
  booktitle = CVPR,
  year = {2023}
}

@inproceedings{ye2023vhoi,
  title={Diffusion-Guided Reconstruction of Everyday Hand-Object Interaction Clips},
  author={Ye, Yufei and Hebbar, Poorvi and Gupta, Abhinav and Tulsiani, Shubham},
  booktitle=ICCV,
  pages={19717--19728},
  year={2023}
}

@inproceedings{swamy2023showme,
  title={SHOWMe: Benchmarking Object-agnostic Hand-Object 3D Reconstruction},
  author={Swamy, Anilkumar and Leroy, Vincent and Weinzaepfel, Philippe and Baradel, Fabien and Galaaoui, Salma and Br{\'e}gier, Romain and Armando, Matthieu and Franco, Jean-Sebastien and Rogez, Gr{\'e}gory},
  booktitle=ICCVW,
  pages={1935--1944},
  year={2023}
}

@Manual{blender,
   title = {Blender - a 3D modelling and rendering package},
   author = {Blender Online Community},
   organization = {Blender Foundation},
   address = {Stichting Blender Foundation, Amsterdam},
   year = {2018},
   url = {http://www.blender.org},
 }

@ARTICLE{Bullock2013HandCentric,
  author={Bullock, Ian M. and Ma, Raymond R. and Dollar, Aaron M.},
  journal={IEEE Transactions on Haptics}, 
  title={A Hand-Centric Classification of Human and Robot Dexterous Manipulation}, 
  year={2013},
  volume={6},
  number={2},
  pages={129-144},
}

@ARTICLE{Cutkosky1989OnGraspChoice,
  author={Cutkosky, M.R.},
  journal={IEEE Transactions on Robotics and Automation}, 
  title={On grasp choice, grasp models, and the design of hands for manufacturing tasks}, 
  year={1989},
  volume={5},
  number={3},
  pages={269-279},
}

@inproceedings{Hodan2020Epos,
  title={{EPOS: Estimating 6d pose of objects with symmetries}},
  author={Hodan, Tomas and Barath, Daniel and Matas, Jiri},
  booktitle=CVPR,
  pages={11703-11712},
  year={2020}
}

@inproceedings{krull2015learning,
  title={Learning analysis-by-synthesis for 6D pose estimation in RGB-D images},
  author={Krull, Alexander and Brachmann, Eric and Michel, Frank and Yang, Michael Ying and Gumhold, Stefan and Rother, Carsten},
  booktitle={Proceedings of the IEEE international conference on computer vision},
  pages={954--962},
  year={2015}
}

@inproceedings{pavlakos2024reconstructing,
  title={Reconstructing hands in 3d with transformers},
  author={Pavlakos, Georgios and Shan, Dandan and Radosavovic, Ilija and Kanazawa, Angjoo and Fouhey, David and Malik, Jitendra},
  booktitle=CVPR,
  pages={9826--9836},
  year={2024}
}

@inproceedings{lin2021end-to-end,
author = {Lin, Kevin and Wang, Lijuan and Liu, Zicheng},
title = {End-to-End Human Pose and Mesh Reconstruction with Transformers},
booktitle = CVPR,
year = {2021},
}

@inproceedings{fan2024hold,
  title={{HOLD}: Category-agnostic 3d reconstruction of interacting hands and objects from video},
  author={Fan, Zicong and Parelli, Maria and Kadoglou, Maria Eleni and Kocabas, Muhammed and Chen, Xu and Black, Michael J and Hilliges, Otmar},
  booktitle=CVPR,
  pages={494--504},
  year={2024}
}

@inproceedings{hampali2023inhand,
  author    = {Hampali, Shreyas and Hodan, Tomas and Tran, Luan and Ma, Lingni and Keskin, Cem and Lepetit, Vincent},
  title     = {In-Hand 3D Object Scanning from an RGB Sequence},
  booktitle   = CVPR,
  year      = {2023},
}

@inproceedings{banerjee2025hot3d,
  title={Hot3d: Hand and object tracking in 3d from egocentric multi-view videos},
  author={Banerjee, Prithviraj and Shkodrani, Sindi and Moulon, Pierre and Hampali, Shreyas and Han, Shangchen and Zhang, Fan and Zhang, Linguang and Fountain, Jade and Miller, Edward and Basol, Selen and others},
  booktitle=CVPR,
  pages={7061--7071},
  year={2025}
}

@inproceedings{ornek2024foundpose,
  author    = {{\"O}rnek, Evin P{\i}nar and Labb\'e, Yann and Tekin, Bugra and Ma, Lingni and Keskin, Cem and Forster, Christian and Hoda{\v{n}}, Tom{\'a}{\v{s}}},
  title     = {FoundPose: Unseen Object Pose Estimation with Foundation Features}, 
  booktitle   = ECCV,
  year      = {2024},
}

@inproceedings{hodavn2020bop,
  title={{BOP challenge 2020 on 6D object localization}},
  author={Hoda{\v{n}}, Tom{\'a}{\v{s}} and Sundermeyer, Martin and Drost, Bertram and Labb{\'e}, Yann and Brachmann, Eric and Michel, Frank and Rother, Carsten and Matas, Ji{\v{r}}{\'\i}},
  booktitle=ECCVW,
  pages={577--594},
  year={2020},
  organization={Springer}
}

@inproceedings{yi2024egoallo,
  title={Estimating body and hand motion in an ego-sensed world},
  author={Yi, Brent and Ye, Vickie and Zheng, Maya and Li, Yunqi and M{\"u}ller, Lea and Pavlakos, Georgios and Ma, Yi and Malik, Jitendra and Kanazawa, Angjoo},
  booktitle={Proceedings of the Computer Vision and Pattern Recognition Conference},
  pages={7072--7084},
  year={2025}
}

@inproceedings{Prakash2024Hands,
    author = {Prakash, Aditya and Tu, Ruisen and Chang, Matthew and Gupta, Saurabh},
    title = {3D Hand Pose Estimation in Everyday Egocentric Images},
    booktitle = ECCV,
    year = {2024}
}

@inproceedings{Prakash2024HOI,
    author = {Prakash, Aditya and Chang, Matthew and Jin, Matthew and Tu, Ruisen and Gupta, Saurabh},
    title = {3D Reconstruction of Objects in Hands without Real World 3D Supervision},
    booktitle = ECCV,
    year = {2024}
}

@inproceedings{dong2024hambasingleview3dhand,
  title={Hamba: Single-view 3d hand reconstruction with graph-guided bi-scanning mamba},
  author={Dong, Haoye and Chharia, Aviral and Gou, Wenbo and Vicente Carrasco, Francisco and De la Torre, Fernando D},
  booktitle=NeurIPS,
  volume={37},
  pages={2127--2160},
  year={2024}
}

@inproceedings{potamias2025wilor,
  title={Wilor: End-to-end 3d hand localization and reconstruction in-the-wild},
  author={Potamias, Rolandos Alexandros and Zhang, Jinglei and Deng, Jiankang and Zafeiriou, Stefanos},
  booktitle=CVPR,
  pages={12242--12254},
  year={2025}
}

@inproceedings{zhou2024simple,
  title={A simple baseline for efficient hand mesh reconstruction},
  author={Zhou, Zhishan and Zhou, Shihao and Lv, Zhi and Zou, Minqiang and Tang, Yao and Liang, Jiajun},
  booktitle=CVPR,
  pages={1367--1376},
  year={2024}
}

@inproceedings{li2024hhmr,
  title={{HHMR: Holistic Hand Mesh Recovery by Enhancing the Multimodal Controllability of Graph Diffusion Models}},
  author={Li, Mengcheng and Zhang, Hongwen and Zhang, Yuxiang and Shao, Ruizhi and Yu, Tao and Liu, Yebin},
  booktitle=CVPR,
  pages={645--654},
  year={2024}
}

@inproceedings{ye2023ghop,
    author = {Ye, Yufei and Gupta, Abhinav and Kitani, Kris and Tulsiani, Shubham},    
    title = {G-HOP: Generative Hand-Object Prior for Interaction Reconstruction and Grasp Synthesis},
    booktitle = CVPR,
    year = {2024}
}

@inproceedings{lin2023harmonious,
  title={Harmonious feature learning for interactive hand-object pose estimation},
  author={Lin, Zhifeng and Ding, Changxing and Yao, Huan and Kuang, Zengsheng and Huang, Shaoli},
  booktitle={Proceedings of the IEEE/CVF Conference on Computer Vision and Pattern Recognition},
  pages={12989--12998},
  year={2023}
}

@inproceedings{2024graspnet,
    title={Dense Hand-Object(HO) GraspNet with Full Grasping Taxonomy and Dynamics},
    author={Cho, Woojin and Lee, Jihyun and Yi, Minjae and Kim, Minje and Woo, Taeyun and Kim, Donghwan and Ha, Taewook and Lee, Hyokeun and Ryu, Je-Hwan and Woo, Woontack and Kim, Tae-Kyun},
    booktitle=ECCV,
    year={2024}
}

@inproceedings{EPICFields2023,
  title={{EPIC Fields}: {M}arrying {3D} {G}eometry and {V}ideo {U}nderstanding},
  author={Tschernezki, Vadim and Darkhalil, Ahmad and Zhu, Zhifan and Fouhey, David and Larina, Iro and Larlus, Diane and Damen, Dima and Vedaldi, Andrea},
  booktitle   = NeurIPS,
  year      = {2023}
}

@inproceedings{Plizzari2025OSNOM,
title={Spatial Cognition from Egocentric Video: Out of Sight, Not Out of Mind},
author={Plizzari, Chiara and Goel, Shubham and Perrett, Toby and Chalk, Jacob and Kanazawa, Angjoo and Damen, Dima},
booktitle={2025 International Conference on 3D Vision (3DV)},
year={2025}
}

@InProceedings{chen2021monorun,
  title     = {{MonoRUn}: Monocular {3D} Object Detection by Reconstruction and Uncertainty Propagation},
  author    = {Hansheng Chen and Yuyao Huang and Wei Tian and Zhong Gao and Lu Xiong},
  booktitle   = CVPR,
pages        = {10379--10388},
  year      = {2021},
}

@InProceedings{goodwin2022zero,
    author  = {Walter Goodwin and Sagar Vaze and Ioannis Havoutis and Ingmar Posner},
    title   = {Zero-Shot Category-Level Object Pose Estimation},
pages        = {516--532},
volume       = {13699},
    booktitle = ECCV,
    year    = {2022},
}

@inproceedings{liu2021utoshape,
  title={{AutoShape}: {R}eal-Time Shape-Aware Monocular {3D} Object Detection},
  author={Zongdai Liu and Dingfu Zhou and Feixiang Lu and Jin Fang and Liangjun Zhang},
  booktitle=ICCV,
pages        = {15621--15630},
  year={2021},
}

@InProceedings{wang2019normalized,
  title     = {Normalized Object Coordinate Space for Category-Level {6D} Object Pose and Size Estimation},
  author    = {He Wang and Srinath Sridhar and Jingwei Huang and Julien P. C. Valentin and Shuran Song and L. Guibas},
  booktitle   = CVPR,
pages        = {2642--2651},
  year      = {2019},
}

@inproceedings{meshrcnn,
  title   = {Mesh {R-CNN}},
  author  = {Georgia Gkioxari and Jitendra Malik and Justin Johnson},
  year    = {2019},
  pages        = {9784--9794},
  booktitle = ICCV
}

@inproceedings{yu2018posecnn,
    Author = {Xiang, Yu and Schmidt, Tanner and Narayanan, Venkatraman and Fox, Dieter},
    Title = {{PoseCNN}: {A} Convolutional Neural Network for {6D} Object Pose Estimation in Cluttered Scenes},
    booktitle   = RSS,
    Year = {2018}
}

@article{wu2024reconstructing,
  title={Reconstructing Hand-Held Objects in 3D},
  author={Wu, Jane and Pavlakos, Georgios and Gkioxari, Georgia and Malik, Jitendra},
  journal={arXiv preprint arXiv:2404.06507},
  year={2024},
}

@article{oquab2024dinov2,
    title={{DINO}v2: Learning Robust Visual Features without Supervision},
    author={Maxime Oquab and Timoth{\'e}e Darcet and Th{\'e}o Moutakanni and Huy V. Vo and Marc Szafraniec and Vasil Khalidov and Pierre Fernandez and Daniel HAZIZA and Francisco Massa and Alaaeldin El-Nouby and Mido Assran and Nicolas Ballas and Wojciech Galuba and Russell Howes and Po-Yao Huang and Shang-Wen Li and Ishan Misra and Michael Rabbat and Vasu Sharma and Gabriel Synnaeve and Hu Xu and Herve Jegou and Julien Mairal and Patrick Labatut and Armand Joulin and Piotr Bojanowski},
    journal={Transactions on Machine Learning Research},
    issn={2835-8856},
    year={2024},
    url={https://openreview.net/forum?id=a68SUt6zFt},
    note={Featured Certification}
}

@inproceedings{wang2025magichoi,
  title={MagicHOI: Leveraging 3D Priors for Accurate Hand-object Reconstruction from Short Monocular Video Clips},
  author={Wang, Shibo and He, Haonan and Parelli, Maria and Gebhardt, Christoph and Fan, Zicong and Song, Jie},
  booktitle=ICCV,
  pages={5957--5968},
  year={2025}
}

@inproceedings{aytekin2025follow,
  title={Follow my hold: Hand-object interaction reconstruction through geometric guidance},
  author={Aytekin, Ayce Idil and Rhodin, Helge and Dabral, Rishabh and Theobalt, Christian},
  booktitle={Thirteenth International Conference on 3D Vision},
  year={2025}
}

@inproceedings{liu2025easyhoi,
  title={EasyHOI: Unleashing the Power of Large Models for Reconstructing Hand-Object Interactions in the Wild},
  author={Liu, Yumeng and Long, Xiaoxiao and Yang, Zemin and Liu, Yuan and Habermann, Marc and Theobalt, Christian and Ma, Yuexin and Wang, Wenping},
  booktitle=CVPR,
  pages={7037--7047},
  year={2025}
}

@InProceedings{chen2025hort,
author       = {Chen, Zerui and Potamias, Rolandos Alexandros and Chen, Shizhe and Schmid, Cordelia},
title        = {{HORT}: Monocular Hand-held Objects Reconstruction with Transformers},
booktitle    = ICCV,
year         = {2025},
}

@inproceedings{yu2025dynamic,
  title={Dynamic reconstruction of hand-object interaction with distributed force-aware contact representation},
  author={Yu, Zhenjun and Xu, Wenqiang and Xie, Pengfei and Li, Yutong and Anthony, Brian W and Zhang, Zhuorui and Lu, Cewu},
  booktitle=ICCV,
  pages={8590--8599},
  year={2025}
}

@inproceedings{cho2023transformer,
  title={Transformer-Based Unified Recognition of Two Hands Manipulating Objects},
  author={Cho, Hoseong and Kim, Chanwoo and Kim, Jihyeon and Lee, Seongyeong and Ismayilzada, Elkhan and Baek, Seungryul},
  booktitle=CVPR,
  pages={4769--4778},
  year={2023}
}

@inproceedings{ismayilzada2025qort,
  title={QORT-Former: Query-optimized real-time Transformer for understanding two hands manipulating objects},
  author={Ismayilzada, Elkhan and Sayem, MD Khalequzzaman Chowdhury and Tiruneh, Yihalem Yimolal and Chowdhury, Mubarrat Tajoar and Boboev, Muhammadjon and Baek, Seungryul},
  booktitle=AAAI,
  volume={39},
  number={4},
  pages={3895--3903},
  year={2025}
}

@misc{wang2024hocapcapturedataset3d,
      title={HO-Cap: A Capture System and Dataset for 3D Reconstruction and Pose Tracking of Hand-Object Interaction},
      author={Jikai Wang and Qifan Zhang and Yu-Wei Chao and Bowen Wen and Xiaohu Guo and Yu Xiang},
      year={2024},
      eprint={2406.06843},
      archivePrefix={arXiv},
      primaryClass={cs.CV},
      url={https://arxiv.org/abs/2406.06843},
}

@inproceedings{jiang2025hand,
  title={Hand-held Object Reconstruction from RGB Video with Dynamic Interaction},
  author={Jiang, Shijian and Ye, Qi and Xie, Rengan and Huo, Yuchi and Chen, Jiming},
  booktitle={Proceedings of the Computer Vision and Pattern Recognition Conference},
  pages={12220--12230},
  year={2025}
}

@article{chen2025sam,
  title={Sam 3d: 3dfy anything in images},
  author={Chen, Xingyu and Chu, Fu-Jen and Gleize, Pierre and Liang, Kevin J and Sax, Alexander and Tang, Hao and Wang, Weiyao and Guo, Michelle and Hardin, Thibaut and Li, Xiang and others},
  journal={arXiv preprint arXiv:2511.16624},
  year={2025}
}
}

\clearpage
\setcounter{section}{0}
\renewcommand{\thesection}{\Alph{section}}
\maketitlesupplementary

\section{Overview}
On the project's webpage, we provide  qualitative videos showcasing qualitative results and include details describing the video in~\cref{sec:qualitative_video}.
Rest of this document is arranged as follows. \Cref{sec:dataset} provides additional annotation details of the EPIC-\timeline and HOT3D-\timeline datasets.
We ablate the robustness of \meth to the boundaries of segments in \timeline in Sec~\ref{sec:ablation-noisy}.
Results on \StableGrasp in the ARCTIC dataset~\cite{fan2023arctic} are provided in~\cref{sec:results}. Additional implementation details are provided in~\cref{suppl:implementation}.
We then qualitatively evaluate CAD-agnostic models in~\cref{sec:cad_agnostic_in_the_wild}.
Finally, in~\cref{sec:limitations}, we discuss limitations of our work.

\section{Qualitative Video}
\label{sec:qualitative_video}

We include videos showcasing the reconstruction results on the two datasets using our proposed approach \meth.
The video collection contains examples from both EPIC-\timeline and HOT3D-\timeline.
In each case, we show the original video (left), 
projected reconstruction in camera frame (middle) 
and 3D hand-object reconstruction from 2 different views (right).
We also show the object and hands in world coordinate frame (bottom) with camera pose as a red prism.

Additionally, we provide examples of \StableGrasp sequences in EPIC-\timeline.
There are two examples from each object category (bottle, can, mug, glass, bowl, cup, plate, pan, saucepan).

\begin{figure}
    \centering
    \includegraphics[width=\linewidth]{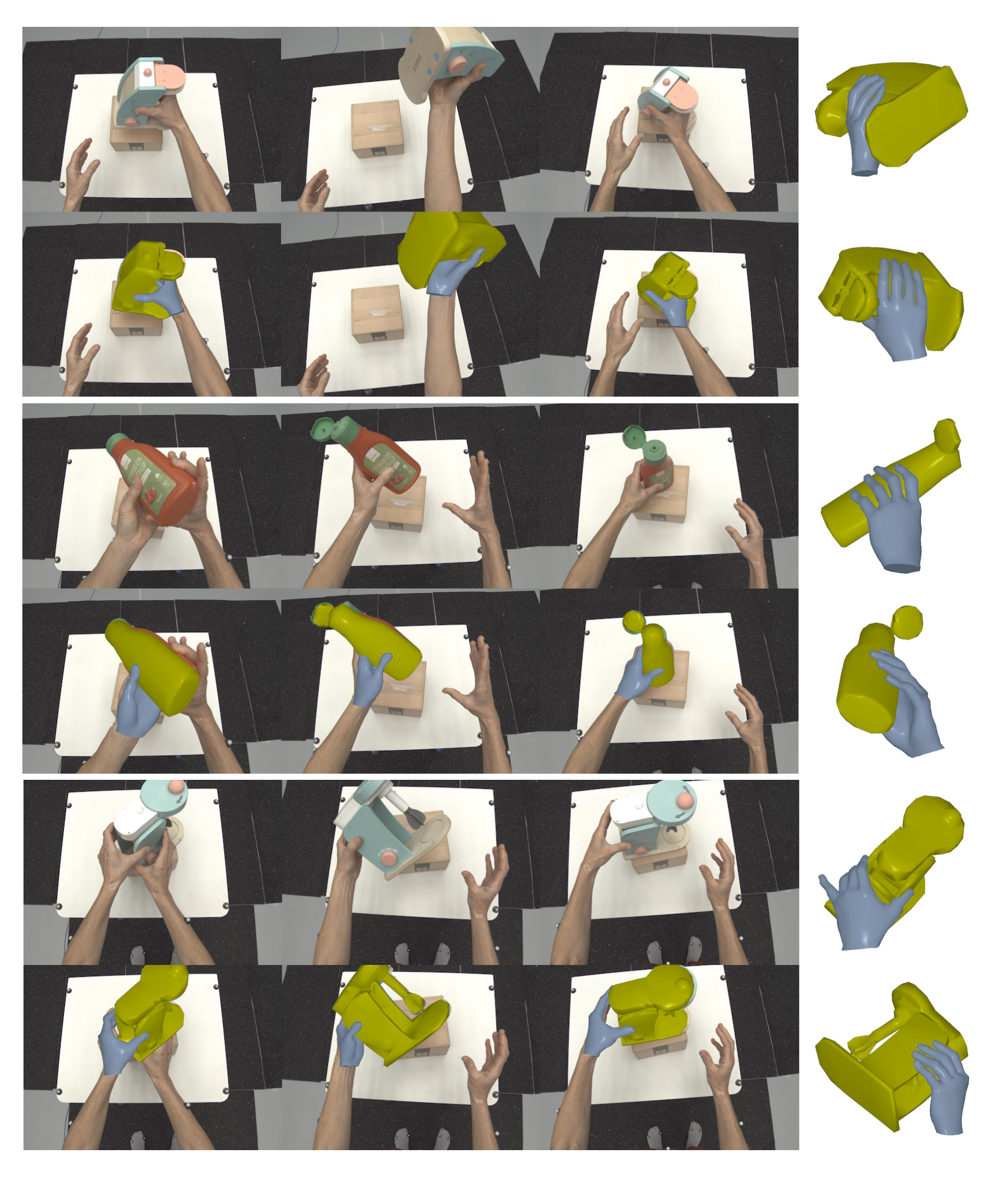}
    \caption{Qualitative results of \meth on \StableGrasp from ARCTIC~\cite{fan2023arctic}.
    There are three sequences visualised here.
    Top row in each sequence contains input frames.
    Bottom row in each sequence contains frames with reconstructed hand and object.
    Last column shows the hand and object reconstruction from two different perspectives.}
    \label{fig:arctic_qual}
\end{figure}

\section{Annotating HOT3D-\timeline and EPIC-\timeline}
\label{sec:dataset}

With the definition of \timeline and \StableGrasp in Section 3.1 of the main paper, we annotate Hand-Interaction Timelines in two datasets.

\subsection{HOT3D-\timeline.}
For the HOT3D~\cite{banerjee2025hot3d} dataset which has 3D ground truth, we automatically extract stable grasps sequences with threshold $\tau = 0.5$ in Equation 1 in the main paper.
We locate $1,239$ stable grasps sequences which we then extend automatically to \timeline using the annotations to identify when the object is in-view. 
In total, we label $113$ HITs covering $410,650$ frames across $20$ videos, $3,288$ segments ($872$ \Static, $1239$ \StableGrasp, $1177$ \UnstableContact) and $22$ objects.

\subsection{EPIC-\timeline.}
We annotate the temporal segments of \timeline from the EPIC-KITCHENS~\cite{Damen2022RESCALING} videos.
This offers a dataset distinct from prior works, which are collected in lab settings 
\cite{BrahmbhattContactPose:Pose, TaheriGRAB:Objects, Zhang2021ManipNet:Representation} or 
contain recordings specifically collected to evaluate grasps with no underlying action \cite{HampaliHOnnotate:Poses, kwon2021h2o, ChaoDexYCB:Objects}.
Instead, we aim to leverage \StableGrasp definition to identify \timeline sequences within unscripted egocentric videos of daily actions.
Note that we exclude interactions with non-rigid objects and only focus on interactions with rigid known objects.
We next detail our annotation pipeline:

\noindent \textbf{1. Identifying candidate clips.} The ultimate goal of hand-object reconstruction is to generalize to any rigid or dynamic objects, including those belonging to novel classes. 
However, as we show later in~\cref{sec:cad_agnostic_in_the_wild}, current approaches for reconstruction of unknown objects~\cite{fan2024hold, ye2023vhoi,Sucar2020NodeSLAM:Reconstruction,Huang2022ReconstructingVideo,swamy2023showme} are still in their infancy. 
We thus restrict our scope to known object categories and focus instead of high-fidelity hand-object reconstruction.
Note that this is distinct from assuming instance-level CAD models -- the general CAD model of a bottle might not exactly match all bottles in daily life.
We exclude tiny objects and shortlist $9$ categories frequently used in kitchens: plate, bowl, bottle, cup, mug, can, pan, saucepan, glass\footnote{For \textbf{object mesh}, we made per-category CAD model in Blender~\cite{blender}.}.
We use annotations and narrations to find clips where a hand is in contact with one of these categories.

\begin{table}[t]
    \caption{Sensitivity to noisy boundary on HOT3D stable grasp subset.}
    \centering
\resizebox{\linewidth}{!}{
    \begin{tabular}{l|cc}
    \toprule
         Method & ADD & SCA-ADD \\
         \midrule
         HOMAN~\cite{Hasson2021TowardsVideos} & 15.0 & 10.0  \\
         \hline
         COP & 70.0 & 32.9 \\
         COP w/ Noisy Stable Grasp Boundaries & 60.0 & 27.4 \\
         \bottomrule
    \end{tabular}
    }
    \label{tab:noisy_boundary}
\end{table}

\begin{figure}[t]
    \centering
    \includegraphics[width=0.2\linewidth]{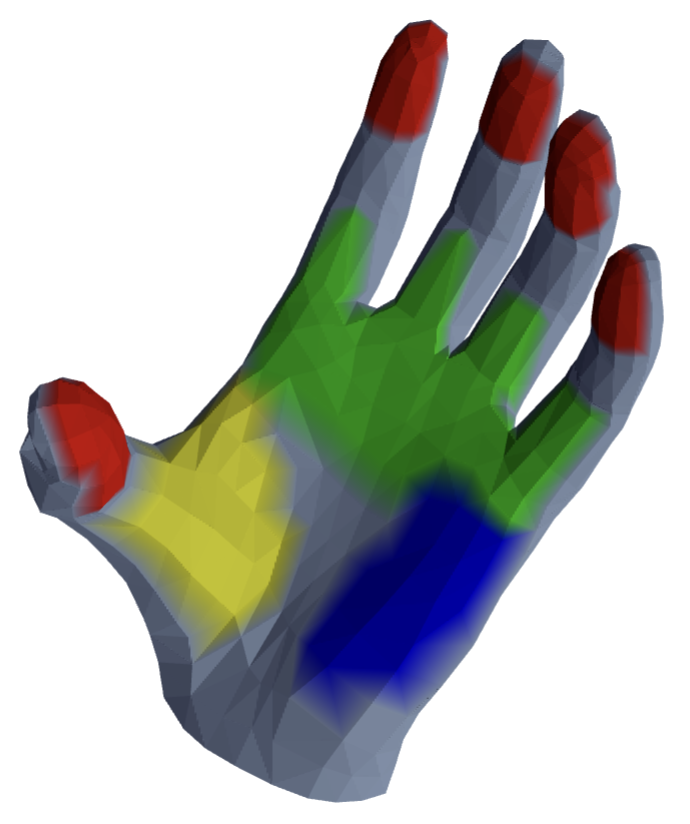}
    \caption{
    Eight contact regions: five fingertips $V_F$ + three palm areas. The contact regions serve two purposes: bounding the object inside and attracting the object closer to these regions.}
    \label{fig:handregions}
\end{figure}

\noindent \textbf{2. Annotating \StableGrasp.}
Two annotators were asked to label the start-and-end frames following the \StableGrasp definition.
We discard segments when, (i)~both the hand and object are out-of-view during the sequence or, (ii)~the object does not match the category CAD model specified.

In total, we label $2,431$ video clips of stable gasps from $141$ distinct videos in $31$ kitchens~\cite{Damen2022RESCALING}.
For each clip, we provide a start and end time of the stable grasp, as well as $319,661$ segmentation masks for the hand and the object during the stable grasp from the dense VISOR annotations~\cite{VISOR2022}.
Of these, $1,446$ contain left hand stable grasps and $985$ contain right hand stable grasps.

\noindent \textbf{3. Annotating \timeline segments.}
Once we have the stable grasps annotated, we extend them to \timeline.
We select $42$ videos that have verified camera pose estimates from~\cite{EPICFields2023} with metric scale and gravity available from~\cite{Plizzari2025OSNOM}.
Manual annotations for temporal segments are then added to form consecutive segments labelled with segment type.
In total, we label $96$ HITs, covering $79,736$ frames and $269$ segments ($135$ \Static, $106$ \StableGrasp, $28$ \UnstableContact).

\begin{table*}[t]
\centering
    \caption{\textbf{Dataset Comparison}. Here we compare various characteristics and labels provided by various datasets. We also show statistics of \StableGrasp and \timeline (when available). $^*$: object poses or segments are not provided. $^\dagger$: subjects in the released train/val set}
   
\resizebox{\textwidth}{!}{%
\begin{tabular}{@{}lrcccccccccccccccc@{}}\toprule
\multirow{3}{*}{Dataset} & \multirow{3}{*}{Year} &\multicolumn{3}{c}{\textbf{Characteristics}} &\multicolumn{3}{c}{\textbf{Labels}} &\multicolumn{5}{c}{\textbf{Stable Grasps' Stats}} &\multicolumn{4}{c}{\textbf{\timeline's Stats}} \\
\cmidrule(lr){3-5}\cmidrule(lr){6-8}\cmidrule(lr){9-13}\cmidrule(lr){14-17}
& & \small{In-the-wild} & \parbox{0.7cm}{\small{Funct.}\\Intent} & Ego & Pose GT & \parbox{1cm}{Stable \\ Grasp} & \parbox{0.6cm}{\timeline} & \#Env & \#Sub & \#Cat & \#Inst & \#Seq & \parbox{1cm}{Avg. \\ Duration} & \#frames & \parbox{1.5cm}{Avg. Seq. \\ Per HIT} & \#Seq \\
 \hline
 FPHA~\cite{Garcia-HernandoFirst-PersonAnnotations} &2018 & \xmark & \cmark & \cmark & 3D & \xmark & \xmark & 3 & 6 & 4 & 4 & 1,175 & - & - & - & - \\

HO3D~\cite{HampaliHOnnotate:Poses} &2020  & \xmark & \xmark & \xmark & 3D & \cmark (part) & \xmark  & 1 & 10 & 10 & 10 & 65 & - & - & - & - \\
ContactPose~\cite{BrahmbhattContactPose:Pose} &2020 & \xmark & \cmark & \xmark & 3D & \cmark & \xmark & 1 & 50 & 25 & 25 & 2,306 & - & - & - & - \\
GRAB~\cite{TaheriGRAB:Objects} &2020 & \xmark & \cmark & \xmark & 3D & \xmark & \xmark & 1 & 10 & 51 & 51 & 1,334 & - & - & - & - \\
 
H2O~\cite{kwon2021h2o} &2021 & \xmark & \cmark & \cmark & 3D & \xmark & \xmark & 3 & 4 & 8 & 8 & 24 & - & - & - & - \\
DexYCB~\cite{ChaoDexYCB:Objects} &2021 & \xmark & \xmark & \xmark & 3D & \xmark & \xmark & 1 & 10 & 20 & 20 & 1,000 & - & - & - & -\\
 
 HOI4D~\cite{Liu2022HOI4D:Interaction} &2022 & \xmark & \cmark & \cmark & 3D & \xmark & \xmark & 610 & 9 & 20 & 800 & 5,000 & - & - & - & - \\
Assembly101~\cite{SenerAssembly101:Activities} &2022 & \xmark & \cmark &\xmark & 3D Hand$^*$ & \xmark & \xmark & 1 & 53 & 15 & 15 & 4,321 & - & - & - & - \\
OakInk~\cite{YangOakInk:Interaction} &2022 & \xmark & \cmark & \xmark & 3D & \xmark & \xmark & 1 & 12 & 32 & 100 & 1,356 & - & - & - & - \\
  SHOWMe~\cite{swamy2023showme} & 2023 & \xmark & \xmark & \xmark & 3D & \cmark & \xmark & 1 & 15 & 42 & 42 & 96 & - & - & - & - \\
  ARCTIC~\cite{fan2023arctic} &2023 & \xmark & \cmark & \cmark & 3D & \xmark & \xmark & 1 & 9$^\dagger$ & 11 & 11 & 339 & - & - & - & - \\
 \textbf{ARCTIC w/ \StableGrasp} &2025 & \xmark & \cmark & \cmark & 3D & \cmark & \xmark & 1 & 9 & 11 & 11 & 1,303 & - & - & - & - \\
  HOGraspNet~\cite{2024graspnet} &2024 & \xmark & \xmark & \cmark & 3D & \cmark & \xmark & 1 & 99 & 30 & 30 & $\sim$3861 & - & - & - & - \\
  HO-Cap~\cite{wang2024hocapcapturedataset3d} & 2024 & \xmark & \xmark & \cmark & 3D & \xmark & \xmark & 1 & 9 & 64 & 64 & 64 & - & - & - & - \\
  HOT3D~\cite{banerjee2025hot3d} &2024 & \xmark & \cmark & \cmark & 3D & \xmark & \xmark & 4 & 19 & 33 & 33 & 295 & - & - & - & - \\
 \textbf{HOT3D-\timeline} (ours) &2025 & \xmark & \cmark & \cmark & 3D & \cmark & \cmark & 4 & 9 & 22 & 22 & 1,239 & 121.1s & 410,650 & 29.1 & 113 \\
 \midrule
Core50~\cite{LomonacoCORe50:Recognition} &2017 & \cmark & \xmark & \xmark & 2D Mask & \xmark & \xmark & 11 & - & 10 & 50 & 550 & - & - & - & - \\
 MOW~\cite{CaoReconstructingWild, Patel2022LearningVideos} &2021 & \cmark & \cmark & \xmark & \xmark & \xmark & \xmark & 500 & 500 & 121 & 500 & 500 & - & - & - & -\\ 
 \textbf{EPIC-\timeline} (ours) &2025 & \cmark & \cmark & \cmark & 2D Mask & \cmark & \cmark & 141 & 31 & 9 & $\sim$390 & 2,431 & 13.8s & 79,736 & 2.8 & 96 \\
 \bottomrule
\end{tabular}
}
    \label{tab:datasetCompareFull}
\end{table*}

\subsection{Dataset Comparison}
\Cref{tab:datasetCompareFull} provides a more comprehensive comparison of our datasets with regularly used datasets for hand-object reconstruction. This is an extension of Table \textcolor{red}{1} in the main paper.

\begin{table*}[t]
    \caption{\textbf{Results on ARCTIC.} 
    \colorbox{green!25}{Green} shows the \colorbox{green!25}{best} performing method per metric and \colorbox{yellow!25}{yellow} shows the \colorbox{yellow!25}{second} best. $\mathrm{COP}^{\dagger}$ is COP without propagation.}
    \centering
    \resizebox{0.95\textwidth}{!}{%
    \begin{tabular}{@{}lcccccccccccc}
\toprule
\multirow{2}{*}{Category} & \multicolumn{4}{c}{SCA-IOU} & \multicolumn{4}{c}{ADD} & \multicolumn{4}{c}{SCA-ADD} \\
\cmidrule(lr){2-5}\cmidrule(lr){6-9}\cmidrule(lr){10-13}
 & HOMan & Rigid~\cite{swamy2023showme} & Dynamic & $\text{COP}^{\dagger}$ & HOMan & Rigid~\cite{swamy2023showme} & Dynamic & $\text{COP}^{\dagger}$ & HOMan & Rigid~\cite{swamy2023showme} & Dynamic & $\text{COP}^{\dagger}$ \\
\midrule
box              & 30.9  & \cellcolor{yellow!25}67.3 & 47.5  & \cellcolor{green!25}71.8 & 33.3  & 37.7  & \cellcolor{yellow!25}52.9 & \cellcolor{green!25}60.1 & 16.9  & \cellcolor{yellow!25}28.7 & 26.7  & \cellcolor{green!25}45.5 \\
capsulemachine   & 34.4  & 34.9                   & \cellcolor{yellow!25}43.8 & \cellcolor{green!25}66.6 & 42.1  & \cellcolor{yellow!25}48.4 & 45.3  & \cellcolor{green!25}57.9 & 24.5  & \cellcolor{yellow!25}34.1 & 26.1  & \cellcolor{green!25}43.7 \\
espressomachine  & 36.9  & 49.2                   & \cellcolor{yellow!25}52.2 & \cellcolor{green!25}73.0 & 44.6  & 48.5  & \cellcolor{yellow!25}64.4 & \cellcolor{green!25}72.3 & 28.8  & 36.2  & \cellcolor{yellow!25}38.2 & \cellcolor{green!25}55.6 \\
ketchup          & 18.0  & 32.7                   & \cellcolor{yellow!25}48.4 & \cellcolor{green!25}62.3 & 15.1  & \cellcolor{yellow!25}51.9 & 39.6  & \cellcolor{green!25}56.6 & 9.3   & \cellcolor{yellow!25}37.8 & 23.7  & \cellcolor{green!25}39.2 \\
laptop           & 35.3  & \cellcolor{yellow!25}62.0 & 51.8  & \cellcolor{green!25}69.1 & 43.8  & 45.1  & \cellcolor{yellow!25}60.4 & \cellcolor{green!25}63.9 & 28.0  & \cellcolor{yellow!25}37.1 & 34.8  & \cellcolor{green!25}46.9 \\
microwave        & 36.1  & \cellcolor{yellow!25}51.5 & 48.8  & \cellcolor{green!25}76.1 & 56.2  & 50.9  & \cellcolor{yellow!25}77.7 & \cellcolor{green!25}83.9 & 27.3  & 35.3  & \cellcolor{yellow!25}41.8 & \cellcolor{green!25}64.2 \\
mixer            & 34.3  & 37.9                   & \cellcolor{yellow!25}51.1 & \cellcolor{green!25}69.4 & 45.1  & 48.4  & \cellcolor{yellow!25}66.4 & \cellcolor{green!25}73.8 & 26.7  & 32.2  & \cellcolor{yellow!25}38.8 & \cellcolor{green!25}54.2 \\
notebook         & 38.7  & \cellcolor{yellow!25}57.8 & 55.4  & \cellcolor{green!25}66.6 & 33.8  & 43.0  & \cellcolor{yellow!25}53.6 & \cellcolor{green!25}61.6 & 20.8  & \cellcolor{yellow!25}33.2 & 32.4  & \cellcolor{green!25}42.7 \\
phone            & 39.5  & 36.7                   & \cellcolor{yellow!25}52.4 & \cellcolor{green!25}62.2 & 28.1  & \cellcolor{yellow!25}39.7 & 34.9  & \cellcolor{green!25}46.6 & 19.9  & \cellcolor{yellow!25}29.2 & 21.4  & \cellcolor{green!25}30.4 \\
scissors         & 5.2   & 0.0                    & \cellcolor{yellow!25}16.2 & \cellcolor{green!25}23.7 & 7.0   & \cellcolor{yellow!25}47.4 & 43.9  & \cellcolor{green!25}70.2 & 5.2   & \cellcolor{yellow!25}36.3 & 25.6  & \cellcolor{green!25}47.7 \\
waffleiron       & 36.3  & 48.3                   & \cellcolor{yellow!25}51.0 & \cellcolor{green!25}65.0 & 45.0  & 59.5  & \cellcolor{yellow!25}72.5 & \cellcolor{green!25}76.3 & 26.2  & 39.2  & \cellcolor{yellow!25}40.1 & \cellcolor{green!25}51.5 \\
\midrule
\textit{Average} & 33.1  & 46.6                   & \cellcolor{yellow!25}49.0 & \cellcolor{green!25}66.1 & 37.1  & 46.9  & \cellcolor{yellow!25}56.0 & \cellcolor{green!25}65.1 & 22.0  & \cellcolor{yellow!25}34.2 & 32.0  & \cellcolor{green!25}46.9 \\
\bottomrule
\end{tabular}

    }
    \label{tab:arctic_grasps_results}
\end{table*}

\begin{table}[t]
    \caption{Ablation on the \StableGrasp Loss $E_{SG}$ variants on ARCTIC. We show improvement over the Dynamic Baseline}
    \centering
\resizebox{\linewidth}{!}{
    \begin{tabular}{cccc}
    \toprule
    Obj. Vert. Selection & Frames Selection & ADD & SCA-ADD \\
    \midrule
    \rowcolor{blue!10} $V_o$ & $N^2$ pairs      & 65.1 (+9.1) & 46.9 \\
    $v^*_o$ & $N^2$ pairs                       & 58.4 (+2.4) & 37.2 \\
    $V_o$ & $N$ consecutive                     & 59.1 (+3.1) & 38.1 \\
    \hline
    \multicolumn{2}{c}{Dynamic Baseline}        & 56.0 (+0.0) & 32.0 \\
    \bottomrule
    \end{tabular}    
    }
    \label{tab:ablate_E_SG_variants}
\end{table}

\begin{table}[t]
    \caption{\textbf{Ablation on the weights.}
    We highlight our choice of $\lambda_1$ and $\lambda_2$ (blue) on ARCTIC}
    \centering
\resizebox{0.6\linewidth}{!}{
    \begin{tabular}{cccc}
    \toprule
    $\lambda_1$& $\lambda_2$ & ADD & SCA-ADD \\
    \midrule
        0 & 0.1 & 56.0 & 32.0 \\
        1 & 0 & 63.0 & 45.0 \\
        \rowcolor{blue!10} 1 & 0.1 & 65.0 & 47.0 \\
        10 & 0.1 & 49.0 & 39.0 \\
    \bottomrule
    \end{tabular}
    }
    \label{tab:ablate_weights}
\end{table}

\section{Dependency on Accurate Boundaries}
\label{sec:ablation-noisy}
COP relies on the provided \timeline segment boundaries.
One limitation of the method is the need for accurate start-end times of all segments in the hit.
These annotations can be relieved if segments are estimated through a localisation model or VLM given a labelled training dataset.
While our results in the paper use labelled segments of \timelinefull (\timeline), we provide an ablation on the need for accurate segment boundaries.

To assess the sensitivity of \meth to labelling boundary accuracy, we add random noise---sampled from $\mathrm{Uniform}(10,\,30)$ frames---to the ground-truth boundaries for 40 randomly selected \StableGrasp samples from HOT3D.
As shown in \Cref{tab:noisy_boundary}, noisy boundaries leads to a performance drop for COP; however, even with such noise, COP still outperforms the baseline~\cite{Hasson2021TowardsVideos} by a large margin.

\section{Sensitivity to Hand Pose Hoise}

We analyse our method's sensitivity to the hand pose noise,
using a random subset of 100 stable grasp segments from HOT3D.
We run HaMeR~\cite{pavlakos2024reconstructing} to obtain the finger poses and use these as input to our method.

\begin{figure}[t]
    \centering
    \includegraphics[width=1\linewidth]{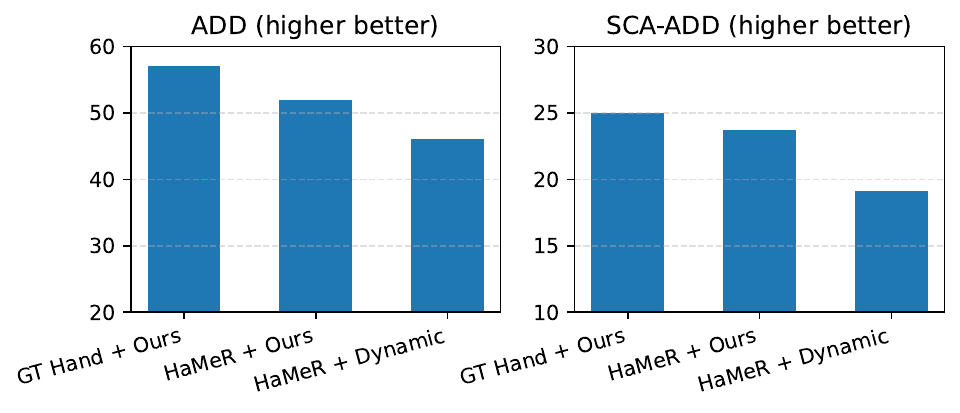}
    \caption{Robustness to noisy hand poses.
    }
    \label{fig:noisy_hand}
\end{figure}

Figure~\ref{fig:noisy_hand} compares the results.
When we switch from ground-truth to estimated hand poses,
our method drops reasonably for both ADD and SCA-ADD metrics.
However, our method still clearly outperforms the \emph{best} performing baseline -- Dynamic.
This finding is in line with the results in the paper.

\section{Results on \StableGrasp in ARCTIC}
\label{sec:results}

In addition to HOT3D~\cite{banerjee2025hot3d} and EPIC-KITCHENS~\cite{DamenScalingDataset}, we also explore the ARCTIC dataset~\cite{fan2023arctic} with 3D ground truth for \timeline reconstruction.
However, due to short clips in the dataset, we only evaluate the stable grasp segments on this dataset.
Similar to HOT3D, we automatically extract stable grasp sequences with threshold of $\tau = 0.5$ and identify $1303$ stable grasp sequences across $9$ subjects covering $11$ categories.
\Cref{tab:arctic_grasps_results} contains per-category results for stable grasps in ARCTIC.
\meth outperforms the baseline~\cite{Hasson2021TowardsVideos} and alternate assumptions on all the $11$ CAD-model categories.
Categories like ``capsule machine" see significant improvement in ADD score (+$12.6$).
On average, \meth improves ADD from $56.0$ with dynamic assumption to $65.1$ using the stable grasp assumption.

\Cref{fig:arctic_qual} shows qualitative results on \StableGrasp from ARCTIC.
In \cref{tab:ablate_E_SG_variants} similar to the ablation in the main paper for HOT3D-\timeline, we ablate \StableGrasp Loss $E_{SG}$ and show improvement over the Dynamic baseline.
Furthermore, in \cref{tab:ablate_weights}, we ablate the weights on ARCTIC and draw similar conclusion as the analogous ablation on the HOT3D dataset (Table~\textcolor{red}{7} in the main paper).

\section{Additional Implementation Details}
\label{suppl:implementation}

\noindent \textbf{Physical Loss $E_{push}$ and $E_{pull}$.}
In the main paper, we note our usage of physical repulsion and attraction losses $E_{push}$ and $E_{pull}$.
These are similar to the repulsion and attraction losses in~\cite{HassonLearningObjects}.

The term $E_{push}$ ensures  all object vertices are located inside 
the contact surface of the hand (\cref{fig:handregions}).
$E_{push}$ applies independently to each frame, hence we omit the superscript $^n$.
For each $v_o \in V_o$, 
we locate the nearest vertex in hand contact regions,
and compute the distance along the surface normal of this hand vertex.
Object vertices that penetrate into the contact surface will have negative values.
We maximise those negative values, truncating the positive ones:
\begin{align}
\label{eq:inside}
E_{push} &= \sum_{v_o \in V_o} -1 * \min( d_v, 0 )\\
d_v &:= \langle v_o - v^*_h, n^*_h \rangle 
\label{eq:inside_dv}
\end{align}
where $v^{*}_h$ is the corresponding nearest vertex on the hand and $n^{*}_h$ is the surface normal of $v^{*}_h$.

In addition to $E_{push}$, which pushes the object out of the penetrating region against the hand, 
we use a balancing loss $E_{pull}$ which pulls the object to touch 
the fingers.
$E_{pull}$ also applies independently to each frame and we omit the superscript $^n$.
We here focus on the contact regions
showcased in Fig~\ref{fig:handregions}.
For each finger tip contact region with hand vertices $\{v_h\}_C$, the region-to-object distance is defined as the minimum distance of all $(v_h, v_o)$ pairs. We use 5 finger tip regions and minimise the average of these region-to-object distances.
\begin{align}
E_{pull} & = \frac{1}{5}\sum_{C}d(\{v_h\}_C, V_o) \\
d(\{v_h\}_C, V_o) & := \min_{v_h \in \{v_h\}_C, v_o \in V_o} \langle v_h - v_o, n_o \rangle
\label{eq:close}
\end{align}
where $n_o$ is the surface normal of $v_o$.

\noindent \textbf{Pose initialisation for \Static segments.}
As the object is typically supported by a surface when static, we use 10 initialisations all with an \textit{upright} orientation.
The initialisations differ in the object's rotation around the axis of support.

\noindent \textbf{Pose initialisation for \StableGrasp segments.}
When using datasets with 3D ground truth, the initial rotations are generated by clustering the ground-truth rotations, where clustering is performed via the axis-angle representation of the rotation matrix. The initial translation is generated by averaging the ground-truth translations. We initialise 10 rotations and 1 global translation for each (object, left/right hand) pair.
For EPIC-HIT, we manually set initial object relative poses to the common poses of each category. 
Each (category, left/right hand) pair has on average 4.1, minimum 1 and maximum 8 initialisation poses.
Importantly, all compared methods (HOMan~\cite{Hasson2021TowardsVideos}, Rigid, Dynamic, COP) are initialised with these same set of initial poses,
ensuring fair comparison in Table-2 and Table-3 in the main paper and Table-3 in the supp.

\noindent \textbf{Pose initialisation for \UnstableContact segments.} We use random initialisation.

\noindent \textbf{Computational cost analysis.}
The main computation is due to mesh projection in $E_{mask}$; $E_{SG}$ is lightweight for meshes with $\approx 500$ vertices.
The Hand-Interaction Timeline (HIT) propagation is inherently sequential, potential speed-up can be gains through engineering the per-segment optimisation, e.g. multiple initialisation in the same segment can be optimised in parallel.

\begin{figure}
    \centering
    \includegraphics[width=\linewidth]{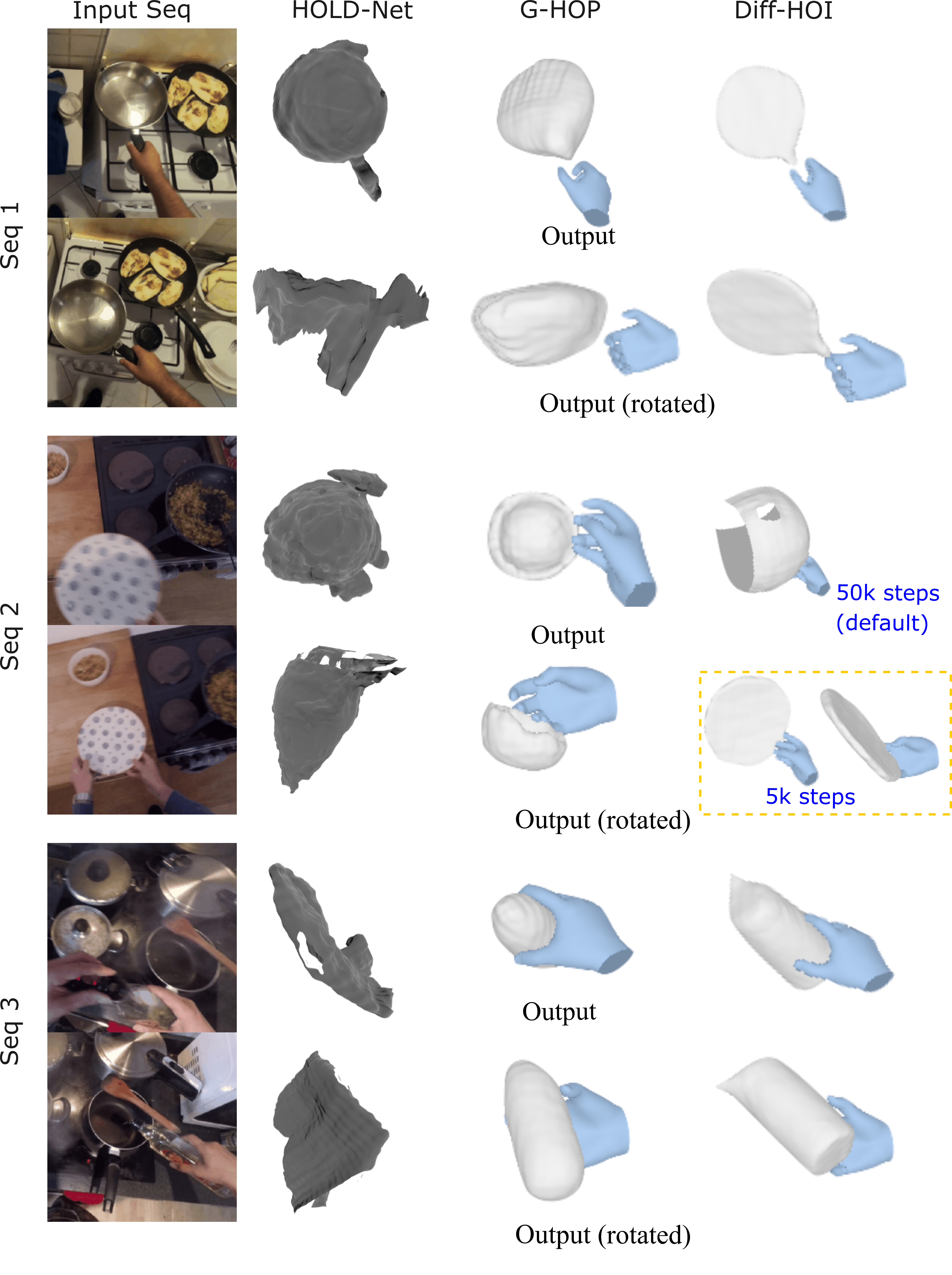}
    \caption{\textbf{In-the-wild qualitative evaluation of~\cite{fan2024hold,ye2023vhoi,ye2023ghop}}. Owing to high occlusion due to fingers, the CAD-agnostic methods struggle to reconstruct the object shapes.}
    \label{fig:wild_examples_full}
\end{figure}

\section{In-the-wild evaluation of CAD-Agnostic methods}
\label{sec:cad_agnostic_in_the_wild}

In our method, we assume knowledge of the CAD model.
We explore works that attempt reconstruction without CAD model's knowledge.
In this section, we showcase these models to be unusable for in-the-wild hand-object reconstruction.

We evaluate CAD-agnostic methods HOLD-Net~\cite{fan2024hold}, G-HOP~\cite{ye2023ghop} and Diff-HOI~\cite{ye2023vhoi} on the \StableGrasp from EPIC-HIT dataset.
HOLD-Net is a neural rendering based multiple-view method, while G-HOP and Diff-HOI are data-driven methods that learn implicit shape priors from in-the-lab datasets.

\Cref{fig:wild_examples_full} shows HOLD-Net is able to reconstruct the object's visible surface. 
However, HOLD-Net is unable to generate the complete object surface due to finger occlusion. 
As input views are typically limited in egocentric videos, HOLD-Net also struggles with the unseen surfaces -- the bottle's symmetry is not reconstructed, see the rotated output.
In-the-wild videos are also challenging for data-driven methods\footnote{authors of these papers acknowledge their limitations in in-the-wild}.
In~\cref{fig:wild_examples_full},  G-HOP fails to produce the shape for the pan and generates a bowl shape for the plate. Diff-HOI also performs poorly. Diff-HOI can generate the plate shape at an intermediate step (see 5K steps result in yellow square), but produces a wrong shape eventually (at the default 50k steps), highlighting robustness limitations.

Overall, these methods are at an infancy stage. Our method can extend to CAD-Agnostic methods when these are more robust.
Importantly, it is not obvious how to quantitatively compare these methods on the same CAD-based metrics due to the need for alignment of the predicted shapes to the ground-truth CAD-model.
This alignment is not obvious and has a significant impact on the numerical evaluation.

We also compare against FoundPose~\cite{ornek2024foundpose}, 
a CAD-known but training free method. FoundPose use DINOv2~\cite{oquab2024dinov2} to build correspondence between the image and the CAD model. Unlike other object pose estimator, FoundPose does not require training, therefore has the potential of scaling to unseen objects.
In Tab.~\ref{tab:compare_data_driven}, we compare results on a random subset of 40 HOT3D stable grasp sequences.
FoundPose is significantly worse than COP.
Note that FoundPose relies on the texture of the instance CAD model, 
which is not a requirement for our method.

\begin{table}[t]
    \centering
    \resizebox{\linewidth}{!}{%
    \begin{tabular}{l|cccc}
    \toprule
         Method & ADD$\uparrow$ & SCA-ADD$\uparrow$ & $err_{\text{rot}}$ ($^\circ$) $\downarrow$ & $err_{\text{trans}}$ (cm) $\downarrow$ \\
         \midrule
         FoundPose~\cite{ornek2024foundpose} & 5.0 & 1.3 & 78.0 & 33.5 \\
         FoundPose~\cite{ornek2024foundpose} w/o texture & 0.0 & 0.0 & 100.1 & 28.9 \\
         COP (Ours) & \textbf{70.0} & \textbf{32.9} & \textbf{39.2} & \textbf{1.3} \\
         \bottomrule
    \end{tabular}
    }
    \caption{Comparison with data-driven methods. We show avg. rotation and translation errors. 
    }
    \label{tab:compare_data_driven}
\end{table}

\section{Limitations and Future Direction}
\label{sec:limitations}

Whilst results in-the-wild are very promising, our pipeline relies on hand pose estimation as a first stage. Despite the robustness incorporated by the multiple-view joint optimisation, our method fails when the predicted hand poses are incorrect (see main paper Figure \textcolor{red}{8}). 
Our method also struggles with extreme occlusions and ambiguity from limited views.

Another limitation of our approach is its reliance on the knowledge of the category's CAD model. We show in~\cref{sec:cad_agnostic_in_the_wild} that current CAD-agnostic methods~\cite{fan2024hold, ye2023ghop, ye2023vhoi} struggle in-the-wild.
CAD-agnostic reconstruction and generalisation to unknown objects is the ultimate goal, however current approaches do not provide sufficiently representative shapes for hand-object reconstruction where accurate object vertices are required for predicting contact.

\begin{figure}[t]
    \centering
    \includegraphics[width=0.8\linewidth]{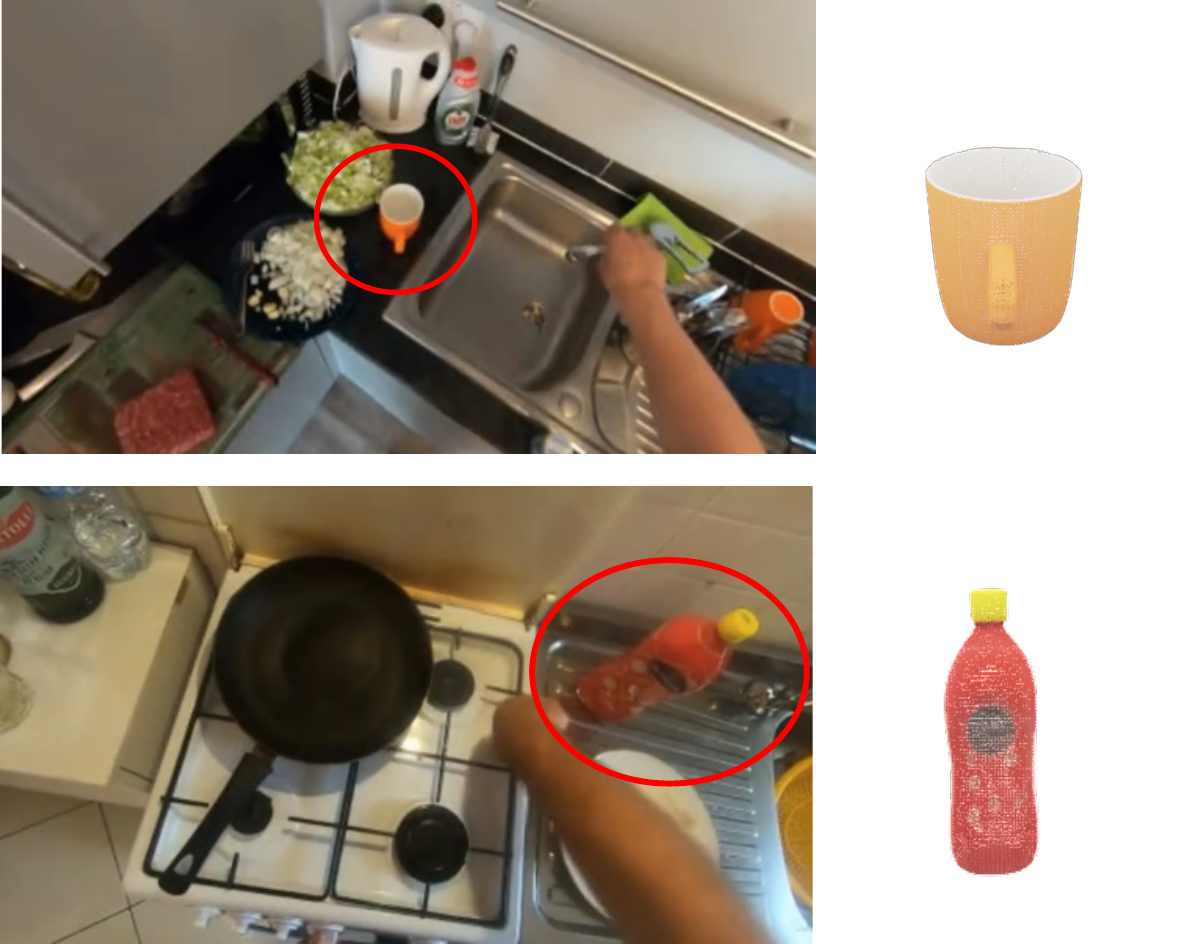}
    \caption{SAM-3D results on EPIC-HIT frames from the Static segments of interaction timelines (i.e. when object is not in contact).
    The predicted models match the CAD models in EPIC-HIT and showcase potential extension into CAD-free assumption.
    }
    \label{fig:sam3d_demo}
\end{figure}

In addition,
we note that the recently published SAM-3D model~\cite{chen2025sam} could be used to obtain candidate CAD models, examples shown in Figure~\ref{fig:sam3d_demo}.
While SAM3D is not integrated into the proposed pipeline,
this is a plausible direction to address the known-CAD limitation.

Finally,
we note that our definition of stable grasp is geometry-based.
Exploring force closure and physical stability is left for future works.

\end{document}